\documentclass[lettersize,journal]{IEEEtran}
\usepackage{amsmath,amsfonts}
\usepackage{algorithmic}
\usepackage{algorithm}
\usepackage{array}
\usepackage[caption=false,font=normalsize,labelfont=sf,textfont=sf]{subfig}
\usepackage{textcomp}
\usepackage{stfloats}
\usepackage{url}
\usepackage{verbatim}
\usepackage{graphicx}
\usepackage{cite}
\usepackage{multirow} 
\usepackage{caption}
\usepackage{graphicx}
\usepackage{array}
\usepackage{import}
\usepackage{colortbl}
\usepackage{booktabs}
\usepackage{xcolor}
\usepackage{orcidlink}
\begin{document}
\title{GenFace: A Large-Scale Fine-Grained Face Forgery Benchmark and Cross Appearance-Edge Learning}
\author{Yaning Zhang\orcidlink{0000-0001-8442-2777}, Zitong Yu\orcidlink{0000-0003-0422-6616}, \emph{Senior Member, IEEE}, Tianyi Wang\orcidlink{0000-0003-2920-6099}, \emph{Member, IEEE}, Xiaobin Huang, Linlin Shen\orcidlink{0000-0003-1420-0815}, \emph{Senior Member, IEEE}, Zan Gao\orcidlink{0000-0003-2182-5741}, \emph{Senior Member, IEEE}, Jianfeng Ren, \emph{Senior Member, IEEE}
\vspace{-2em}
\thanks{This work was supported by the National Natural Science Foundation of China under Grant 82261138629 and 62306061 and 62372325; Guangdong Basic and Applied Basic Research Foundation under Grant 2023A1515010688; Shenzhen Municipal Science and Technology Innovation Council under Grant JCYJ20220531101412030, and open project of National Engineering Laboratory for Big Data System Computing Technology, Shenzhen University, Shenzhen 518060, PR China; Shandong Province National Talents Supporting Program (No.2023GJJLJRC-070); Natural Science Foundation of Tianjin Municipality (No.23JCZDJC00280); and Shandong Project towards the Integration of Education and Industry (No.2022JBZ01-03). (Yaning Zhang and Zitong Yu contributed equally to this work.) (Corresponding author: Linlin Shen and Zan Gao)}

\thanks{Y. Zhang and X. Huang are affiliated with Computer Vision Institute, College of Computer Science and Software Engineering, Shenzhen University, Shenzhen, 518060, China. E-mail: zhangyaning0321@163.com and 2017192014@email.szu.edu.cn }

\thanks{Z. Yu is with School of Computing and Information Technology, Great Bay University, Dongguan, 523000, China, and also with National Engineering Laboratory for Big Data System Computing Technology, Shenzhen University, Shenzhen 518060, China. E-mail: yuzitong@gbu.edu.cn}

\thanks{T. Wang is with Nanyang Technological University, 50 Nanyang Ave, Block N 4, 639798, Singapore. E-mail: terry.ai.wang@gmail.com}

\thanks{L. Shen is with Computer Vision Institute, College of Computer Science and Software Engineering, Shenzhen University, Shenzhen, 518060, China, also with National Engineering Laboratory for Big Data System Computing Technology, Shenzhen University, also with Shenzhen Institute of Artificial Intelligence and Robotics for Society, Shenzhen, 518129 and also with Guangdong Key Laboratory of Intelligent Information Processing, Shenzhen University. E-mail: llshen@szu.edu.cn}

\thanks{Z. Gao is with the Shandong Artificial Intelligence Institute, Qilu University of Technology (Shandong Academy of Sciences), Jinan, 250014, China, and also with the Key Laboratory of Computer Vision and System, Ministry of Education, Tianjin University of Technology, Tianjin, 300384, China. E-mail: zangaonsh4522@gmail.com }

\thanks{J. Ren is with the School of Computer Science, University of Nottingham Ningbo China. E-mail: Jianfeng.Ren@nottingham.edu.cn }
}

\markboth{Journal of \LaTeX\ Class Files,~Vol.~18, No.~9, September~2020}%
{How to Use the IEEEtran \LaTeX \ Templates}

\maketitle
\vspace{-2em}
\begin{abstract}
The rapid advancement of photorealistic generators has reached a critical juncture where the discrepancy between authentic and manipulated images is increasingly indistinguishable. Thus, benchmarking and advancing techniques detecting digital manipulation become an urgent issue. Although there have been a number of publicly available face forgery datasets, the forgery faces are mostly generated using GAN-based synthesis technology, which does not involve the most recent technologies like diffusion. The diversity and quality of images generated by diffusion models have been significantly improved and thus a much more challenging face forgery dataset shall be used to evaluate SOTA forgery detection literature.  In this paper, we propose a large-scale, diverse, and fine-grained high-fidelity dataset, namely GenFace, to facilitate the advancement of deepfake detection, which contains a large number of forgery faces generated by advanced generators such as the diffusion-based model and more detailed labels about the manipulation approaches and adopted generators. In addition to evaluating SOTA approaches on our benchmark, we design an innovative Cross Appearance-Edge Learning (CAEL) detector to capture multi-grained appearance and edge global representations, and detect discriminative and general forgery traces. Moreover, we devise an Appearance-Edge Cross-Attention (AECA) module to explore the various integrations across two domains. Extensive experiment results and visualizations show that our detection model outperforms the state of the arts on different settings like cross-generator, cross-forgery, and cross-dataset evaluations. Code and datasets will be available at \url{https://github.com/Jenine-321/GenFace}.
\end{abstract}


\section{Introduction}
\vspace{-0.5em}
 \begin{table}[!ht]
 	\centering
 			\caption{An overview of face forgery datasets. IQA: Image quality assessment. FG: Fine-grained; FF++: FaceForensics++; DFDC: Deepfake detection challenge; DF-1.0: Deeperforensics; DFFD: Diverse fake face dataset. GFW: Generated faces in the wild. We randomly select some images and feed them to the pre-trained facial Image Quality Assessment (IQA) model, ResNet50 \cite{faceqnet}, to evaluate image quality. Higher IQA scores mean higher image quality.}
 	\setlength{\tabcolsep}{0.00004mm}{
 		\resizebox{0.49\textwidth}{!}{%
 			\normalsize
 			\begin{tabular}{@{\hspace{0.001pt}}cccccc@{\hspace{-1em}}ccc@{\hspace{0.8em}}c@{\hspace{1em}}c}
 				\toprule
 				\multirow{2}[4]{*}{Dataset} & \multicolumn{2}{c}{Generator} &
 				\multicolumn{1}{c}{\multirow{2}[4]{*}{\shortstack{Attribute \\ Editing}}}&
 				\multicolumn{1}{c}{\multirow{2}[4]{*}{\shortstack{ FG \\Label}}}& \multicolumn{1}{c}{\multirow{2}[4]{*}{\shortstack{\hspace{0.8em}Public\\ \hspace{0.8em}Avalibility}}} & \multicolumn{1}{c}{\multirow{2}[4]{*}{\shortstack{Real\\ Images}} } & \multicolumn{1}{c}{\multirow{2}[4]{*}{\shortstack{ Fake\\Images}}} &
 				\multicolumn{1}{c}{\multirow{2}[4]{*}{\shortstack{Image \\Resolution}}}&
 				\multicolumn{1}{c}{\multirow{2}[4]{*}{\shortstack{\hspace{-0.8em}IQA \\\hspace{-0.8em}Score}}}&		
 				\multicolumn{1}{c}{\multirow{2}[4]{*}{\shortstack{Year}}} \\
 				\cmidrule{2-3}      & Diff. & GAN   &       &       &       &  \\
 				\midrule
 				UADFV \cite{UADF} & $\times$     & $\checkmark$     & $\times$ & $\times$ &  $\checkmark$   & 241   & 252 & 294$\times$500  &6.13  & 2019 \\
 				Fakespotter \cite{FakeSpotter} & $\times$      &$\checkmark$   & $\checkmark$  & $\times$  & $\times$   & 6, 000 & 5, 000 & - &- & 2019 \\
 				FF++ \cite{FF++} & $\times$    & $\checkmark$  &$\times$   &$\times$  & $\checkmark$     & - & - & 480$\times$1080 &6.51  & 2019 \\
 				Celeb-DF \cite{Celeb-DF} & $\times$    & $\checkmark$  &$\times$ & $\times$   & $\checkmark$     & - & - &256$\times$256 & 6.64& 2020 \\
 				DFDC \cite{DFDC} & $\times$    & $\checkmark$  &$\times$ &$\times$  & $\checkmark$     & - & - & 256$\times$256 & 6.62& 2020 \\
 				DF-1.0 \cite{DF1.0} & $\times$    & $\checkmark$  &$\times$  & $\times$  & $\checkmark$     & - & - & 1920$\times$1080 &6.93 & 2020 \\
 				DFFD \cite{DFFD}  & $\times$     &$\checkmark$  &$\checkmark$ &$\times$ &$\checkmark$     & 58, 703 & 240, 336 &299$\times$299 &6.95 & 2020 \\
 				ForgeryNet \cite{ForgeryNet} & $\times$    & $\checkmark$  &$\checkmark$ &$\checkmark$    & $\checkmark$     &\hspace{-0.4em}1, 438, 201& \hspace{0.4em} 1, 457, 861 &240\raisebox{-.5ex}{\textasciitilde}1080 & 6.15  & 2021 \\
 			GFW \cite{GFW} &$\checkmark$     &$\checkmark$     &$\times$ &$\times$ &$\checkmark$    & 30, 000 &15, 076 &250\raisebox{-.5ex}{\textasciitilde}512 & 6.96  & 2022 \\
 				Mundra et al. \cite{mundra} &$\checkmark$     &$\times$     &$\times$ &$\times$ &$\checkmark$    & 100, 000 & 41, 500 &500\raisebox{-.5ex}{\textasciitilde}1024 & 6.96 & 2023 \\
 				DiffusionFace \cite{diffusionface} &$\checkmark$     &$\times$     &$\checkmark$ &$\times$ &$\checkmark$    & 30, 000 & 600, 000 &256$\times$256& 6.97 &2024 \\
 				GenFace &$\checkmark$     &$\checkmark$     &$\checkmark$ &$\checkmark$ &$\checkmark$    & 100, 000 & 515, 000 &256\raisebox{-.5ex}{\textasciitilde}1024& 6.98  & 2024 \\
 				\bottomrule
 			\end{tabular}%
 		}
 		\label{tab1}
 	}
 \end{table}%
 
Photorealistic synthesis technologies \cite{HOU2019183,NIPS2014_5ca3e9b1,StyleGene}, especially the latest deep learning-based generative models such as Generative Adversarial Nets (GAN) \cite{NIPS2014_5ca3e9b1,StyleGAN3,StyleGAN2} and Denoising Diffusion Probabilistic Models (DDPM) \cite{Ho2020DenoisingDP,coll}, have made remarkable progress in generating facial fake images, which leads to widespread public concerns about massive malicious abuse of them \cite{Adrian,University2021Deepfakes}. Consequently, benchmarking and advancing facial forgery analysis has become a critical and urgent issue to benefit face forgery detection.

\begin{figure*}[!t]
	\centering
	\includegraphics[width=1\linewidth]{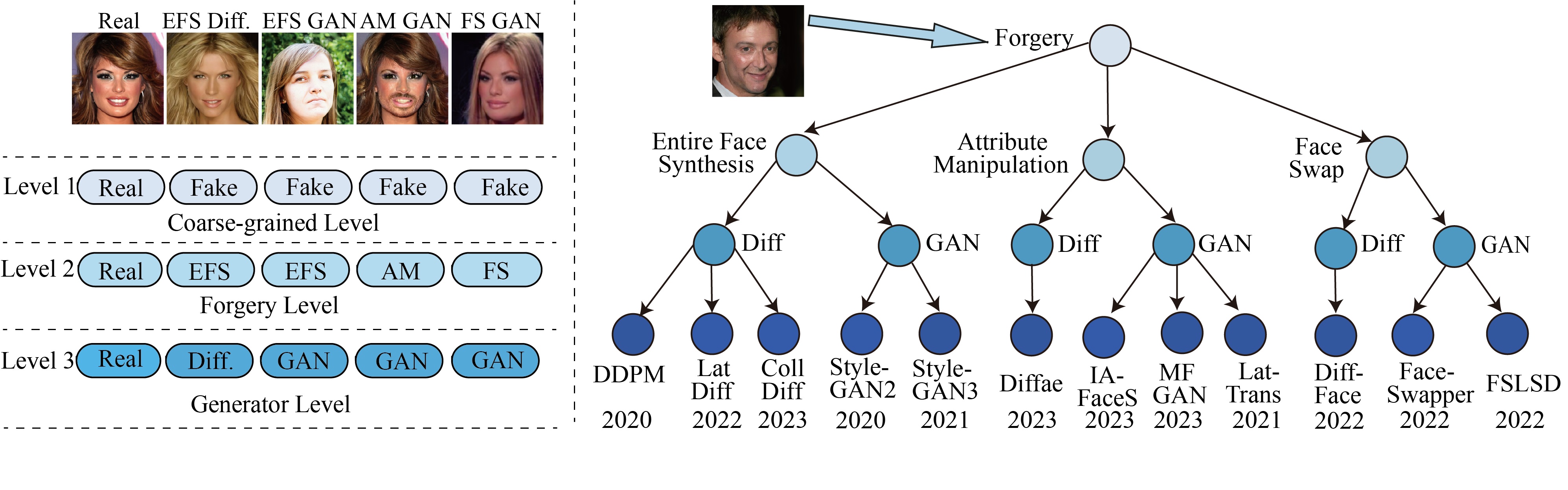}
		\vspace{-3em}
	\centering
	\caption{Taxonomy of the GenFace dataset. At level 1, we divide images into real or fake faces. The second level, i.e., forgery level, classifies forged images into three types, i.e., Entire Face Synthesis (EFS), Attribute Manipulation (AM), and Face Swap (FS). Then, we separate images based on whether forgery approaches are diffusion-based or GAN-based. The final level refers to the specific generators. LatDiff is latent diffusion \cite{lad}. CollDiff is collaborative diffusion \cite{coll}. Diffae is diffusion autoencoders \cite{diffusion}. DiffFace is diffusion face \cite{diff}. MFGAN is MaskFaceGAN \cite{pernuvs2023maskfacegan}. LatTrans is latent transformer\cite{Latent}.}\label{fig1}
\end{figure*}

Face digital manipulation attacks, namely deepfake, generate forgery faces using deep learning-based methods, which mainly involve three categories, i.e., Entire Face Synthesis (EFS) \cite{StyleGAN2,StyleGAN3,Ho2020DenoisingDP}, Attribute Manipulation (AM) \cite{Latent,IA-FaceS,pernuvs2023maskfacegan}, and Face Swap (FS) \cite{FaceSwapper,FSLSD}. As Fig.~\ref{fig1} shows, EFS intends to yield non-existing fake faces from random noises with generative techniques. AM aims to edit facial attributes of an original image to produce a new forgery face with deep leaning-based models. FS proposes to substitute the identity of source images with that of target images using neural networks. The above digital face forgery techniques facilitate the creation of deepfake datasets.  

However, there are the following limitations for existing deepfake datasets: \textbf{(1) Small-scale forgery images}: As Table~\ref{tab1} shows, some early datasets such as UADFV \cite{UADF} and Fakespotter \cite{FakeSpotter} contain fewer than 10,000 forgery images. \textbf{(2) Low image resolution and quality}: Existing efforts \cite{UADF,Celeb-DF,DFDC,DFFD} generate forgery images with resolutions smaller than 1024$\times$1024, and obtain lower Image Quality Assessment (IQA) scores. \textbf{(3) Coarse-grained labels}: Some recent works \cite{Celeb-DF,DFDC,DFFD,DF1.0,FF++} only provide a coarse-grained manipulation label (real or fake). Fine-grained labels are rarely available. \textbf{(4) Poor diversity}: Most works \cite{UADF,Celeb-DF,DFDC,DFFD,DF1.0,FF++,ForgeryNet} tend to synthesize facial images through GAN-based generative networks such as StyleGAN \cite{Stylegan}, while novel generative techniques are hardly considered. To solve the above challenges, inspired by \cite{10205306}, we construct a large-scale, hierarchical, and fine-grained high-fidelity deepfake dataset, namely GenFace, which introduces face forgery images generated by diffusion-based models to provide fine-grained forgery labels. Unlike \cite{10205306} which is applied to image manipulation detection in general natural scenes, our dataset is designed for face forgery detection. Specifically, as displayed in Fig.~\ref{fig1}, we divide deepfakes into three types, i.e., EFS, AM, and FS. We then categorize EFS, AM, and FS into diffusion-based and GAN-based methods, respectively. Each category contains some specific generators, and we provide sample labels at each level. In detail, we generate entire fake face images using StyleGAN2 \cite{StyleGAN2}, StyleGAN3 \cite{StyleGAN3}, DDPM \cite{Ho2020DenoisingDP}, etc., since StyleGAN2 redesigns the normalization layer, and introduces progressive training techniques to gradually increase image resolution, yielding clearer and more detailed images than StyleGAN. However, a large number of image details created by StyleGAN2 appear to be fixed in pixel coordinates, leading to the texture sticking issue. To solve this problem, StyleGAN3 \cite{StyleGAN3} designs the rotation equivariance formula to achieve translation and rotation invariance, which significantly improves the quality of synthesized images. DDPM \cite{Ho2020DenoisingDP} models the conditional probability distribution of pixels, and introduces deformation operations and multi-scale modeling, thereby generating images with better diversity, richness, and realism. Likewise, we create attribute manipulated face images using Diffae \cite{diffusion}, LatTrans \cite{Latent}, etc., and synthesize identity-swapped faces by DiffFace \cite{diff}, FSLSD \cite{FSLSD}, and FaceSwapper \cite{FaceSwapper}. As shown in Table~\ref{tab1}, the IQA scores of GenFace are higher than that of other datasets, since GenFace contains large and diverse high-resolution facial images generated via advanced diffusion generators.

\begin{figure}[!t]%
	\includegraphics[width=\linewidth]{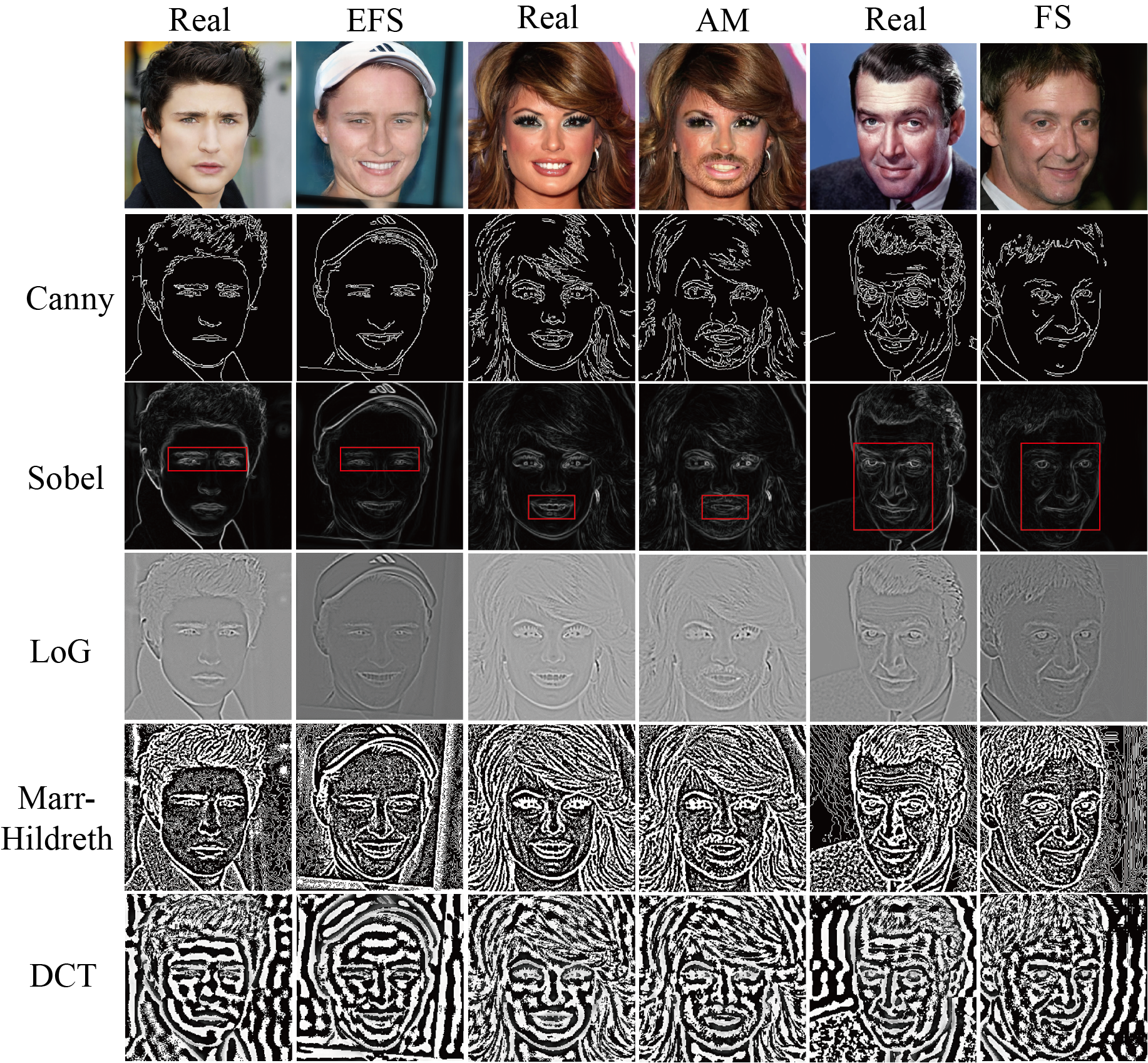}
	\centering
	\caption{The visualization of images produced by different operators. The first row represents the Red Green Blue (RGB) images. Every two columns display the real and fake samples of various manipulations. }\label{fig2}
	\vspace{-1.5em}
\end{figure} 

Large and diverse high-quality deepfake datasets also promote the development of face forgery detection networks. Most detectors \cite{CViT,CEViT,Xception} treat deepfake detection as a binary classification problem, examining manipulated artifacts to discern authenticity using neural networks. To capture comprehensive forgery patterns, some works \cite{CViT,CEViT} leverage hybrid transformer-based networks to capture local and global appearance representations. However, (1) they are not good at capturing high-frequency information in the appearance domain \cite{Inceptiontransformer}. (2) They capture local or global forgery traces only in the appearance domain, limiting their representation ability. In this work, we aim to introduce discriminative features as complementary information, to facilitate deepfake detection. We are motivated by the observation that there are significant differences between authentic and forgery edge images extracted by Sobel \cite{sobel} (see Fig.~\ref{fig2}), compared to other operators such as edge Canny \cite{canny}, Laplacian-of-Gaussian (LoG) \cite{laplacian}, and MarrHildreth \cite{marrh}, as well as the frequency-based Discrete Cosine Transform (DCT) \cite{DCT}. In detail, edge features extracted by the Sobel operator are visually evident in real images, but not obvious in fake ones. Therefore, we utilize the Sobel operator to obtain edge images, and design the edge transformer encoder to explore global forgery traces in the edge domain. Furthermore, inspired by \cite{CEViT}, we design a fine-grained transformer encoder to extract fine-grained appearance global manipulated features. A coarse-grained transformer encoder is proposed to explore coarse-grained global forgery artifacts, to retain high-frequency details in the appearance domain. Moreover, a multi-grained cross-attention module is introduced to explore diverse and comprehensive forgery patterns by integrating multi-grained appearance global embeddings, since counterfeit patterns exist and vary at different image granularities \cite{M2TR}. To achieve better interactions between the appearance and edge domains, we devise an Appearance-Edge Cross-Attention (AECA) module to mine complementary and diverse forgery traces. In summary, the contributions of our work are as follows:

$\bullet$ To the best of our knowledge, we conduct the first non-video, diverse, large-scale, and fine-grained deepfake image dataset, namely GenFace, to facilitate deepfake detection, which contains face forgery images generated by state-of-the-art synthesis techniques like diffusion, and eliminates the need for volunteers to participate in face swap, considerably reducing the cost of data collection.

$\bullet$ We design a novel Cross Appearance-Edge Learning (CAEL) detector, which captures multi-grained appearance and edge global forgery patterns, and explores the diverse fusion across two domains to mine complementary and comprehensive forgery artifacts.

$\bullet$ We conduct a comprehensive benchmarking evaluation using GenFace, demonstrating the effectiveness of the proposed CAEL through cross-generator, cross-forgery, and cross-dataset evaluation. 

\vspace{-1em}

\section{Related Works}\label{sec2}

{\bfseries\setlength\parindent{0em}  Deepfake datasets.} Comprehensive and extensive benchmarks for deepfake detection are limited in the community. Wang et al. \cite{FakeSpotter} created a dataset with entire fake face images generated by the pre-trained ProGAN \cite{celebahq}  and StyleGAN2 \cite{StyleGAN2}. To further increase the diversity of the benchmark, DFFD \cite{DFFD} created the deepfake dataset with facial attribute edited images using face APP, and entire fake face images produced by StarGAN. Video-based face forgery models became available with the release of FaceForensics \cite{FF}, which contains face2face \cite{thies2016face2face} manipulated frames from over 1,000 videos. An enhanced version, FaceForensics++ (FF++) \cite{FF++}, expands the collection to include deepfake \cite{obama2018video} and faceswap \cite{obama2018videoFS} manipulations. Kwon et al. \cite{kodf} created the Korean deepfake (KoDF) dataset by inviting 353 Korean subjects to partake in the crowdsourcing task to participate in face swap. Jiang et al. \cite{DF1.0} built a large-scale dataset for real-world face forgery detection by employing 100 paid actors to record the source videos. By contrast, we utilized the publicly available face datasets CelebAHQ \cite{celebahq} and Flickr-Faces-HQ (FFHQ) \cite{Stylegan} as pristine data, and open-source forgery methods to generate face forgery images, substantially reducing the cost of data collection. However, the above datasets are limited to GAN-based methods such as StyleGAN. For each manipulation type, they hardly take the fine-grained partitioning such as diffusion-based or GAN-based, into account. To address this limitation, we construct the first large-scale, hierarchical, and fine-grained deepfake dataset with diverse and most recent forgery methods, including entire fake face images using DDPM \cite{Ho2020DenoisingDP}, LatDiff \cite{lad}, CollDiff \cite{coll}, StyleGAN2 \cite{StyleGAN2}, and StyleGAN3 \cite{StyleGAN3}, face attribute manipulated images using LatTrans \cite{Latent}, MFGAN\cite{pernuvs2023maskfacegan}, Diffae\cite{diffusion} and IAFaceS \cite{IA-FaceS}, and face-swapped images generated by FaceSwapper \cite{FaceSwapper}, DiffFace\cite{diff} and FSLSD \cite{FSLSD}. 

{\bfseries\setlength\parindent{0em}  Deepfake detection.} Existing deep learning-based models have been proposed to detect the security threat caused by deepfakes. Dang et al. \cite{DFFD} leveraged the Xception backbone to capture local forgery patterns for deepfake detection, which rarely considers global information. To address this problem, Wodajo et al. \cite{CViT} designed a Convolutional Vision Transformer (CViT) that combines CNN with the Vision Transformer (ViT) to identify authenticity. Coccomini et al. \cite{CEViT} studied global forgery traces by integrating EfficientNet \cite{efficientnet} with ViT. To further explore comprehensive forgery patterns, they also encoded multi-scale global features by combining EfficientNet with a CrossViT \cite{CrossViT} for deepfake detection. To enrich embeddings, a Multi-modal Multi-scale Transformer (M2TR) \cite{M2TR} model attempts to introduce high-frequency features extracted by DCT, and fuse them with multi-scale embeddings. Spatial-Frequency Dynamic Graph (SFDG) \cite{SFDG} exploits the relation-aware features in spatial and frequency domains via dynamic graph learning. GramNet \cite{gramnet} leverages global image texture representations for robust fake image detection. Diffusion Reconstruction Error (DIRE) \cite{DIRE} measures the error between an input image and its reconstruction counterpart by a pre-trained diffusion model to distinguish between the real and fake images. UFD \cite{UniversalFakeDetect} utilizes a feature space not explicitly trained to discern authenticity via nearest neighbor and linear probing as instantiations. By contrast, our method is capable of examining global edge forgery traces, as well as multi-grained appearance-edge fusion representations. 

\section{GenFace Construction}\label{tab3}
\begin{figure*}[ht!]%
	\includegraphics[width=\linewidth]{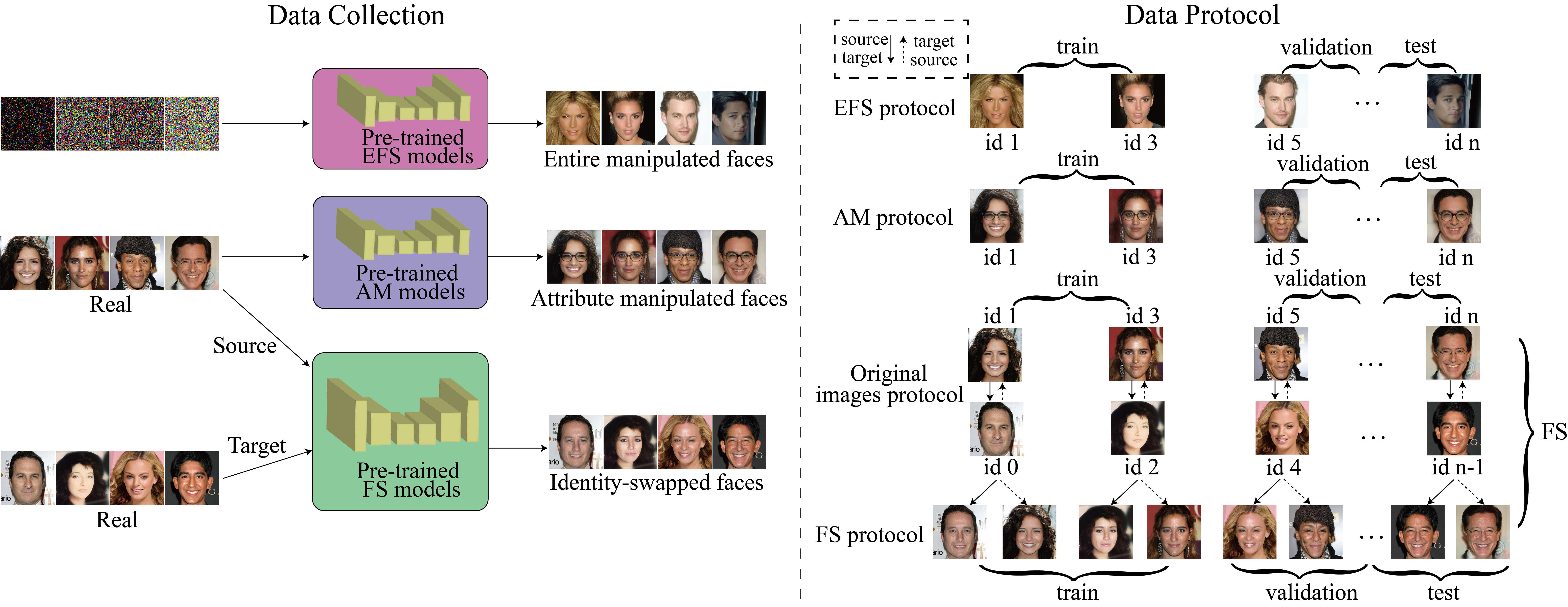}
	\centering
	\caption{Schematic illustration of the collection and partitioning of GenFace. }\label{fig4}
	\vspace{-1em}
\end{figure*}

\begin{table}[t]
	\centering
			\caption{The outline of generator details. Image source means the original dataset that generators are trained on.}
	\label{tab2}}
	\setlength{\tabcolsep}{0.3mm}{
		\resizebox{0.47\textwidth}{!}{
			\normalsize
			\begin{tabular}{ccccc@{\hspace{1em}}c}
				\toprule
				\shortstack{Forgery \\ Method} & \shortstack{ Forgery \\ Type} & \shortstack{Image \\ Source}& \shortstack{Image \\ Number} & \shortstack{Image \\ Resolution} &Year\\
				\midrule
				DDPM \cite{Ho2020DenoisingDP} & EFS   & CelebAHQ & 50k & 256$\times$256& 2020 \\\
				LatDiff \cite{lad} & EFS   & CelebAHQ/FFHQ & 60k &256$\times$256 & 2022 \\
				CollDiff \cite{coll}  & EFS   & CelebAHQ & 50k &512$\times$512 &2023 \\
				StyleGAN2 \cite{StyleGAN2} & EFS   & FFHQ  & 50k &1024$\times$1024 & 2020 \\
				StyleGAN3 \cite{StyleGAN3} & EFS   & CelebAHQ & 50k &256$\times$256  & 2021 \\
				FaceSwapper \cite{FaceSwapper} & FS    & CelebA  & 30k &256$\times$256  & 2022 \\
				Diffface \cite{diff} & FS    & CelebAHQ  & 30k & 1024$\times$1024 &2022 \\
				FSLSD \cite{FSLSD} & FS    & CelebAHQ & 30k &1024$\times$1024& 2022 \\
				LatTrans \cite{Latent} & AM    & CelebAHQ & 60k &1024$\times$1024&  2021 \\
				Diffae \cite{diffusion} & AM    & FFHQ & 70k &256$\times$256& 2023 \\
				IAFaceS \cite{IA-FaceS} & AM    & CelebAHQ & 5k &256$\times$256& 2023 \\
				MFGAN \cite{pernuvs2023maskfacegan} & AM    & CelebAHQ & 30k &1024$\times$1024& 2023 \\			
				\bottomrule
			\end{tabular}%
		}
\vspace{-0.5em}
\end{table}%
\vspace{-1em}
\subsection{Data Collection}\label{subsec31}
{\bfseries\setlength\parindent{0em} Original real images.} CelebAHQ \cite{celebahq} is selected as the original real image dataset, which consists of 30,000 compressed images with a resolution of 1024 × 1024, and is a high-quality version of the CelebA dataset. CelebA \cite{Celeba} includes 10,000 identities, each of which has a collection of twenty images, making a total of 200,000 images. These images are subjected to a series of high-quality processing and image quality assessments, sorted accordingly, and the top 30,000 high-resolution 1024 × 1024 images are retained as the CelebAHQ dataset. We also choose the FFHQ \cite{Stylegan} dataset as the original real image dataset, which is a high-quality face image dataset consisting of 70,000 compressed high-quality images at 1024×1024 resolution, containing considerable variation in age, ethnicity, and image background.

{\bfseries\setlength\parindent{0em} Entire face synthesis.} As Fig.~\ref{fig4} shows, we input random noises into pre-trained diffusion-based or GAN-based models to synthesize fake faces. As Table~\ref{tab2} shows, for diffusion-based methods, we use CelebAHQ pre-trained DDPM \cite{Ho2020DenoisingDP} and LatDiff\cite{lad}, with a sampling step size of 100 and 50 to generate 50,000 and 30,000 images, respectively. The CollDiff \cite{coll} pre-trained by CelebAHQ or FFHQ produces 30,000 images,  respectively, with a sampling step of 100. For GAN-based methods, we leverage StyleGAN2 \cite{StyleGAN2} pre-trained by FFHQ, and StyleGAN3 \cite{StyleGAN3} pre-trained by CelebAHQ, to obtain 50,000 images each. 

{\bfseries\setlength\parindent{0em} Attibute manipulation.} We use facial attribute editing models \cite{Latent,pernuvs2023maskfacegan, IA-FaceS} pre-trained by CelebAHQ to perform attribute manipulations on the original real faces. Since the model was trained using data from the same domain, the quality of the synthesized faces is higher. LatTrans \cite{Latent} can edit forty types of facial attributes, including eyeglasses, bangs, bags, etc. We select some common categories such as eyeglasses and bushy eyebrows. We fed the original real images into LatTrans to generate 30,000 manipulated images for each attribute, respectively. MFGAN \cite{pernuvs2023maskfacegan} can manipulate fourteen local attributes, and we chose the attribute of the big nose to produce 30,000 fake faces. We utilize IAFaceS \cite{IA-FaceS} to manipulate five face attributes of real images, i.e., bags, beard, bushy, open mouth, and narrow eyes, to create 1,000 images for each attribute. For the diffusion-based model, the FFHQ pre-trained Diffae \cite{diffusion} is leveraged to create 70,000 attribute manipulated images.

{\bfseries\setlength\parindent{0em} Face swap.} We leverage the pre-trained face swap model \cite{FaceSwapper,FSLSD} to synthesize fake faces from the source and target image. Specifically, as Fig.~\ref{fig4} shows, we divide the 30,000 original images into two groups, each of which has 15,000 images. One group serves as the source image, and the other group is used as the target object, forming 15,000 source and target image pairs. The identities of image pairs are exchanged to form 15,000 new source and target pairs. Each pair is passed through the face swap model to produce identity-swapped face images. We employ FaceSwapper \cite{FaceSwapper} pretrained on CelebA and FSLSD \cite{FSLSD} pretrained on CelebAHQ, to generate 30,000 fake images each. For the diffusion-based method, we use CelebAHQ pre-trained DiffFace \cite{diff}, with a sampling step size of 1000, to yield 30,000 face-swapped images
\vspace{-1em}
\subsection{Data Protocol}\label{subsec32}
{\bfseries\setlength\parindent{0em} Original real images.} CelebA is officially divided into three parts, i.e., 160,000 images of the ﬁrst 8,000 identities for training, 20,000 images of another 1,000 identities for validation, and 20,000 images of the remaining 1,000 identities for testing. Since identities in CelebAHQ correspond to those in CelebA, we use CelebAHQ along with the official protocols, i.e., 24,183 images for training, 2,993 images for validation, and 2,824 images for testing, to ensure identity independence and exclusivity across all subsets. Similarly, FFHQ is officially split into two parts, i.e., 60,000 images for training, and 10,000 images for testing.

{\bfseries\setlength\parindent{0em} Entire face synthesis.} Note that images from EFS are identity-independent. Therefore, 30,000 samples are selected for each generator of EFS to balance the two categories. We split them in the same ratio as the original image dataset, with 24,183 images in the training set, 2,993 images in the validation set, and 2,824 images in test set. 

{\bfseries\setlength\parindent{0em} Attribute manipulation.} We find that face attribute edited images are matched to the real images in terms of identity. Consequently, we adhere to the official partitioning for each generator, to ensure that the quantity and identity within each subset align with those in the real image dataset.

{\bfseries\setlength\parindent{0em} Face swap.} As Fig.~\ref{fig4} illustrates, the identities of face-swapping images correspond to that of the source images from the CelebAHQ dataset. Therefore, for each generator, we adopt the official protocols to guarantee that the quantity and identity within each subset match those in the real image dataset. 

\begin{figure}[t]%
	
	\includegraphics[width=\linewidth]{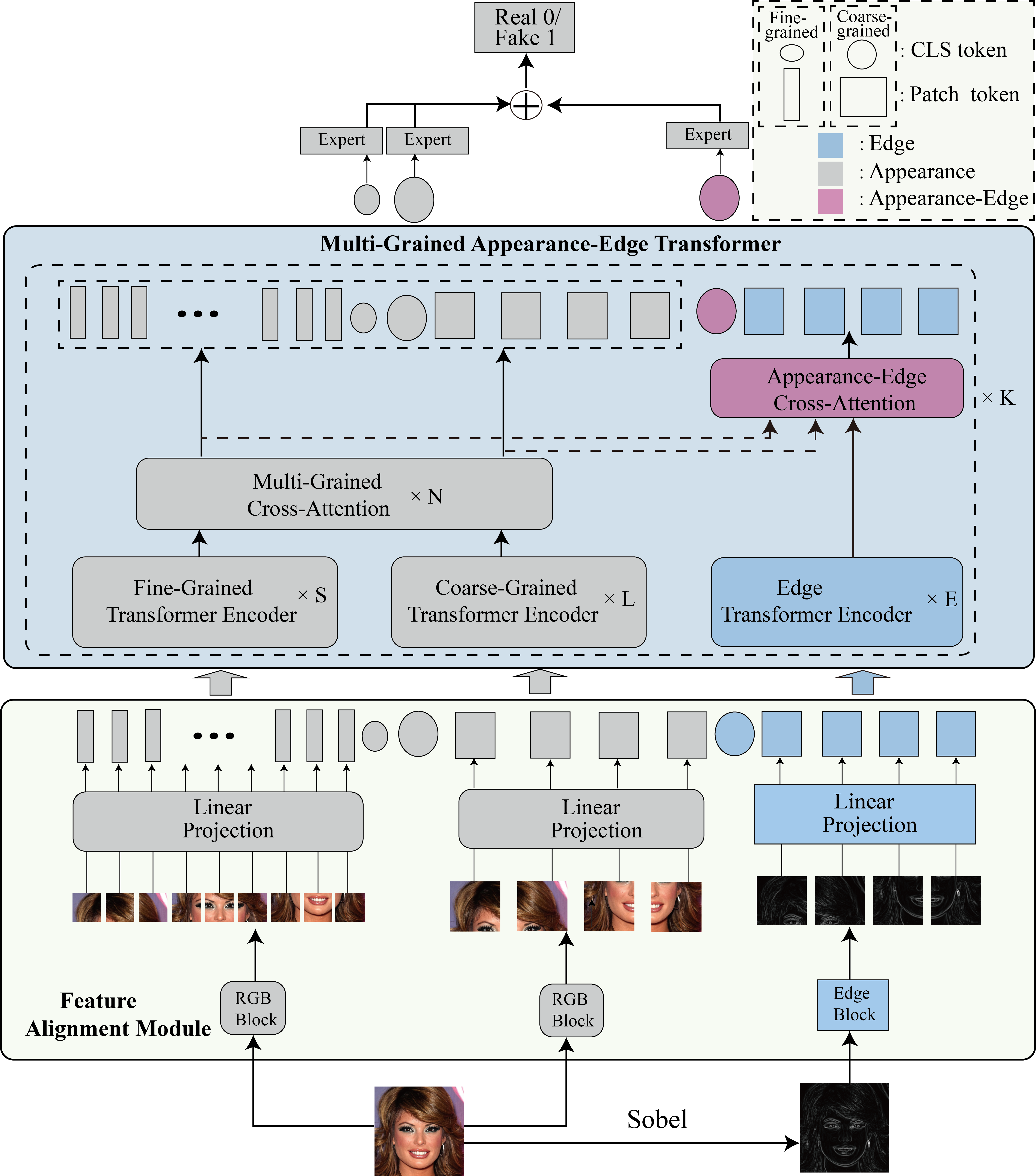}
	\centering
	\caption{The architecture of the proposed model. We first encode fine-grained appearance features, coarse-grained appearance representations, and edge embeddings from input RGB images and edge images via a feature alignment module, respectively. We then fed them into the Multi-grained Appearance-Edge Transformer (MAET) module to capture diverse appearance-edge forgery patterns, global edge features, and diverse mixture representations across two domains. Finally, they are sent to respective experts, each of which consists of a fully connected layer, to yield a single output. The outputs of multiple experts are element-wisely summed and then fed into a softmax function to generate the final prediction.}\label{fig5}
\end{figure}
\begin{figure}[t]%
	\includegraphics[width=\linewidth]{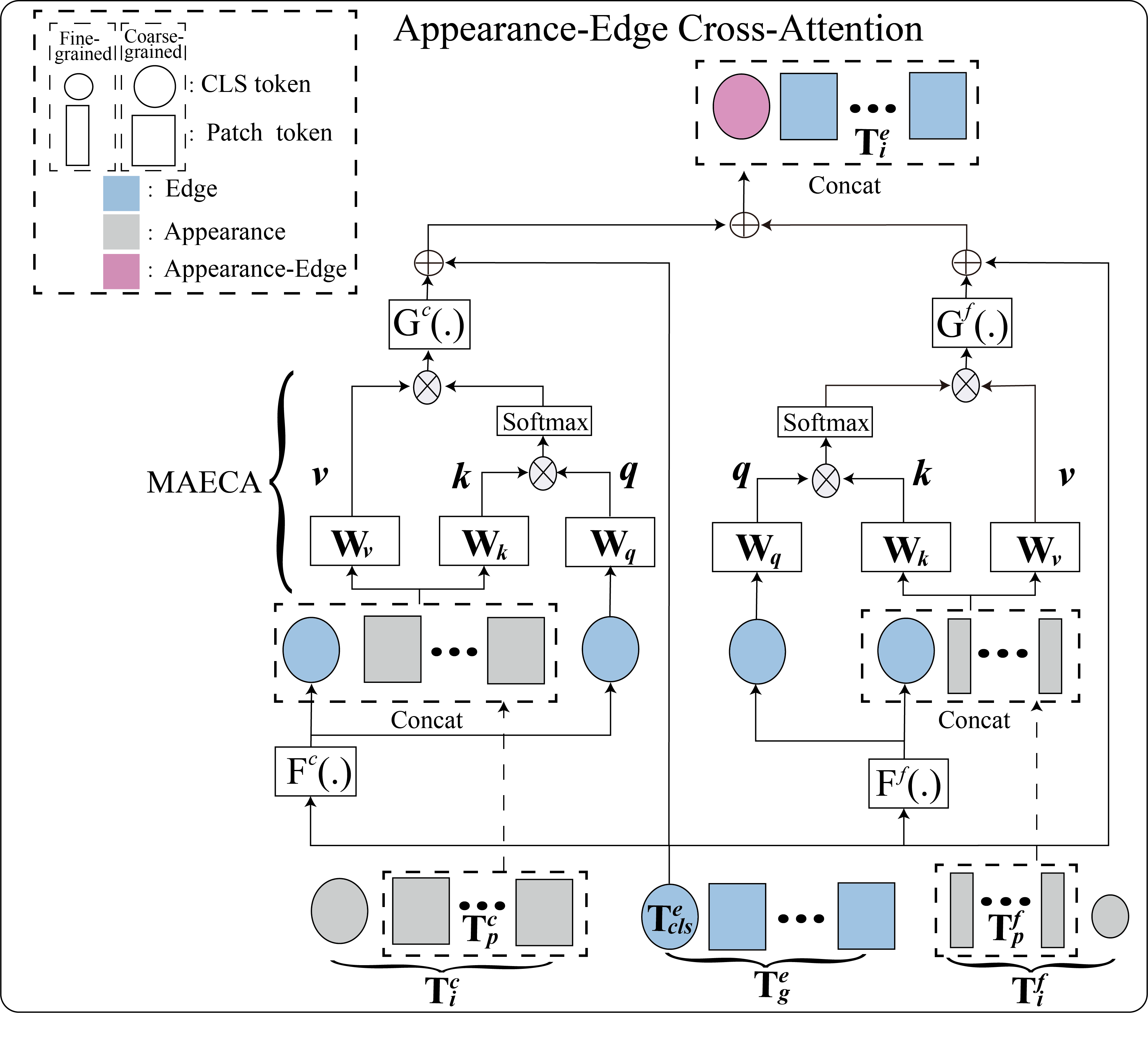}
	\centering
	\caption{The workflow of appearance-edge cross-attention. }\label{fig6}
\end{figure}

\section{Deepfake Detection Networks}\label{sec4}
To capture multi-grained appearance and edge forgery traces, we design Cross Appearance-Edge Learning (CAEL) for deepfake detection. As Fig.~\ref{fig5} shows, CAEL extracts appearance and edge embeddings, respectively, through the feature alignment module (see Sect.~\ref{subsec44}). After that, they are fed into the Multi-Grained Appearance-Edge Transformer (MAET) module (see Sect.~\ref{subsec41}) to capture multi-grained appearance forgery patterns as well as global edge representations, and various synergies across two domains are explored through the AECA module (see Sect.~\ref{subsec42}). Finally, they are fed into respective experts, each of which consists of a fully connected layer, to produce a single output. The outputs of experts are element-wisely summed and then passed through a softmax function to obtain the final prediction.

\subsection{ Feature Alignment Module}\label{subsec44}
Given an input face appearance image $X^a\in\mathbb{R}^{3\times H\times W}$, where $H$ and $W$ denote the height and width, respectively, we utilize the Sobel operator to get the edge image $X^e\in\mathbb{R}^{1\times H\times W}$. To align features in the embedding space, Fig.~\ref{fig5} shows that we send the $X^a$ to two stream RGB blocks with stacked convolutional layers to get the fine-grained features $X^f\in\mathbb{R}^{C\times(\frac{H}{32})\times(\frac{W}{32})}$ and coarse-grained representations $X^c\in\mathbb{R}^{C\times(\frac{H}{4})\times(\frac{W}{4})}$, respectively, where $C$ denotes channels. Meanwhile, the edge image is fed into the edge block with stacked convolutional layers to yield edge local embeddings $X_l^e\in\mathbb{R}^{C\times(\frac{H}{4})\times(\frac{W}{4})}$. To facilitate the extraction of global edge forgery traces, the feature patches in $X_l^e$ are flattened as feature tokens along the channel, and then projected and attached to the class token, to get edge token representations $X_t^e\in\mathbb{R}^{(n+1)\times2d}$, where $n$ and $d$ denote the number and dimension of the feature token, respectively. We can gain fine-grained and coarse-grained appearance token embeddings $X_t^f\in\mathbb{R}^{(n+1)\times d}$ and $X_t^c\in\mathbb{R}^{(n+1)\times2d}$ in the same way, respectively. They are then added with the positional information, and transferred to the first transformer block in the MAET module.

\subsection{Multi-Grained Appearance-Edge Transformer}\label{subsec41}
Unlike traditional transformers \cite{CEViT,CViT} that solely focus on local and global appearance information or multi-grained appearance global forgery patterns, we design the MAET module to encode edge global features and multi-grained appearance global embeddings, and comprehensively integrate them through the AECA module. 

Specifically, as shown in Fig.~\ref{fig5}, MAET mainly contains $K$ transformer blocks, each of which is composed of a fine-grained transformer encoder with $S$ blocks, a coarse-grained transformer encoder with $L$ blocks, an edge transformer encoder with $E$ blocks, a multi-grained cross-attention module with $N$ blocks and an AECA module. The first three transformer encoders adopt the same architecture as ViT. Inspired by \cite{CEViT}, the multi-grained cross-attention module is introduced to mine the multi-grained fused forgery traces.

In detail, given fine-grained appearance tokens $T^f\in\mathbb{R}^{(n+1)\times d}$, coarse-grained appearance tokens $T^c\in\mathbb{R}^{(n+1)\times2d}$, and edge tokens $T^e\in\mathbb{R}^{(n+1)\times2d}$, we send them to the fine-grained transformer encoder, coarse-grained transformer encoder, and edge transformer encoder to obtain global forgery patterns $T_{g}^f\in\mathbb{R}^{(n+1)\times d}$, $T_{g}^c\in\mathbb{R}^{(n+1)\times2d}$, and $T_g^e\in\mathbb{R}^{(n+1)\times2d}$, respectively. To flexibly fuse multi-grained appearance representations, $T_{g}^f$ and $T_{g}^c$ are then fed into the multi-grained cross-attention module to obtain integrated global forgery traces at different granularities $T_{i}^f\in\mathbb{R}^{(n+1)\times d}$ and $T_{i}^c\in\mathbb{R}^{(n+1)\times 2d}$. The detailed workflow of the module is discussed in this work \cite{CrossViT}. To facilitate diverse global interactions between edge and multi-grained appearance features, we design the AECA module to produce $T_i^e\in\mathbb{R}^{(n+1)\times 2d}$ by inputting $T_{i}^f$, $T_{i}^c$, and $T_{g}^e$. That is:
\begin{align}
T_i^e=\textrm{AECA}(T_i^f,\ T_i^c,T_g^e).
\end{align}
The detailed workflow of AECA is discussed in Section~\ref{subsec42}. After that, we transfer $T_{i}^f$, $T_{i}^c$, and ${T}_i^e$ into the subsequent transformer block.

\subsection{ Appearance-Edge Cross-Attention}\label{subsec42}
To fuse the multi-grained appearance-edge representations more efficiently and effectively, we first employ the edge class token as a query to swap information among the fine-grained and coarse-grained appearance patch tokens, respectively, and then back project it to its own branch. Since the edge class token already captures the edge information among all its own patch tokens, interacting with the appearance patch tokens from various granularity helps to fuse appearance and edge information, comprehensively. After the integration across two domains, the edge class token interacts with its own patch tokens again at the next transformer block, where it passes the learned knowledge from different grained appearance tokens to its own edge patch token, to enrich the embedding of each patch token. In detail, as Fig.~\ref{fig6}  illustrates, given fine-grained appearance tokens  $T_{i}^f\in\mathbb{R}^{(n+1)\times d}$, coarse-grained appearance tokens $T_{i}^c\in\mathbb{R}^{(n+1)\times2d}$, and edge tokens $T_{g}^e\in\mathbb{R}^{(n+1)\times2d}$, we can obtain the edge class token $T_{cls}^e\in\mathbb{R}^{1\times2d}$, fine-grained appearance patch tokens $T_p^f\in\mathbb{R}^{n\times d}$, and coarse-grained appearance patch tokens $T_p^c\in\mathbb{R}^{n\times2d}$. The $T_{cls}^e$ serves as a query to interact with $T_p^f$ and $T_p^c$, respectively. The workflow of the fusion between $T_{cls}^e$ and $T_p^f$ is as follows:

Firstly, to boost the information exchange, we project $T_{cls}^e$ to obtain${\ T}_{cls\prime}^e\in\mathbb{R}^{1\times d}$, i.e., ${{\ T}_{cls\prime}^e=\ F}^f(T_{cls}^e)$, and concatenate it with $T_p^f$ to get $T_{all}^{ef}\in\mathbb{R}^{(n+1)\times d}$, where $F^f$ is the projection function for dimension alignment. We then conduct Multi-Head Appearance-Edge Cross-Attention (MAECA) between ${T}_{cls\prime}^e$ and $T_{all}^{ef}$. Mathematically, the calculation of MAECA is as follows:
\begin{align}
	q&={ T}_{cls\prime}^e\ w_q,\\
	k&=T_{all}^{ef}\ w_k,\\
	v&=T_{all}^{ef}\ w_v,
\end{align}
where $w_q,w_k, w_v\in\mathbb{R}^{d\times d_m}$, $m$ denotes the number of head space, and $d_m=d/m $ denotes the dimension of feature tokens in the head space.

Then, the appearance-edge cross-attention weights are calculated as follows:
\vspace{-0.3cm}
\begin{gather}
	A=\textrm{softmax}(\frac{qk^T}{\sqrt{d_m}}),
\end{gather}
where $A\in\mathbb{R}^{1\times(n+1)}$ denotes the attention map between appearance and edge tokens. Note that we only use the class token as a query, the computational cost of producing the attention map in A is linear instead of quadratic like vanilla Multi-Head Self-Attention (MHSA) \cite{transformer}, which makes the entire process more efficient.

After that, to achieve global fusion between edge and fine-grained appearance representations, we design the formula as below:
\vspace{-0.3cm}
\begin{gather}
	\textrm{MAECA}(T_{all}^{ef}) =Av.
\end{gather}

Thereafter, the output of MAECA is projected and added to $T_{cls}^e$ to get $T_{cls}^{ef}\in\mathbb{R}^{1\times 2d}$. 
The interaction between $T_{cls}^e$ and $T_p^c $ follows the same workflow as that between $T_{cls}^e$ and $T_p^f$. Therefore, we can obtain $T_{cls}^{ec}\in\mathbb{R}^{1\times 2d}$ with the fused edge and coarse-grained appearance forgery patterns. Finally, we enhance the communication between $T_{cls}^{ef}$ and $T_{cls}^{ec}$, and then concatenate it with the edge patch tokens $T_p^e$, which can be formulated as: 
\vspace{-0.2cm}
\begin{gather}
	{T}_i^e = [(T_{cls}^{ef}+T_{cls}^{ec})||T_p^e].
\end{gather}

\begin{table}[t]
	\centering
		\caption{Cross-generator evaluation. ACC and AUC scores (\%) of different approaches trained and tested using images produced by diffusion and GAN generators. The best results are in bold.}
	\setlength{\tabcolsep}{3.0mm}{
		\resizebox{0.48\textwidth}{!}{%
			\normalsize
			\begin{tabular}{ccrrrr}
				\toprule
				\multirow{3}[6]{*}{Training Set} & \multirow{3}[6]{*}{Model} & \multicolumn{4}{c}{Testing Set } \\
				\cmidrule{3-6}      &       & \multicolumn{2}{c}{Diffusion-syn.} & \multicolumn{2}{c}{GAN-syn.} \\
				\cmidrule{3-6}      &       & \multicolumn{1}{c}{ACC} & \multicolumn{1}{c}{\cellcolor[gray]{0.9}AUC} & \multicolumn{1}{c}{ACC} & \multicolumn{1}{c}{\cellcolor[gray]{0.9}AUC} \\
				\midrule
				\multirow{11}[6]{*}{Diffusion-syn.} & ResNet \cite{resnet} & \multicolumn{1}{c}{100} & \multicolumn{1}{c}{\cellcolor[gray]{0.9}100} & \multicolumn{1}{c}{50.02 } & \multicolumn{1}{c}{\cellcolor[gray]{0.9}73.92 } \\
				& Xception \cite{Xception} & \multicolumn{1}{c}{100} & \multicolumn{1}{c}{\cellcolor[gray]{0.9}100} & \multicolumn{1}{c}{50.00 } & \multicolumn{1}{c}{\cellcolor[gray]{0.9}58.29} \\
				& VGG \cite{vgg}   & \multicolumn{1}{c}{100} & \multicolumn{1}{c}{\cellcolor[gray]{0.9}100} & \multicolumn{1}{c}{50.00 } & \multicolumn{1}{c}{\cellcolor[gray]{0.9}57.85 } \\
				\cmidrule{2-6}      & ViT \cite{ViT}  & \multicolumn{1}{c}{99.19 } & \multicolumn{1}{c}{\cellcolor[gray]{0.9}99.96 } & \multicolumn{1}{c}{49.88 } & \multicolumn{1}{c}{\cellcolor[gray]{0.9}49.46 } \\
				& CrossViT \cite{CrossViT} & \multicolumn{1}{c}{99.50 } & \multicolumn{1}{c}{\cellcolor[gray]{0.9}99.98 } & \multicolumn{1}{c}{50.14 } & \multicolumn{1}{c}{\cellcolor[gray]{0.9}54.32 } \\
				& iFormer \cite{Inceptiontransformer} & \multicolumn{1}{c}{100} & \multicolumn{1}{c}{\cellcolor[gray]{0.9}100} & \multicolumn{1}{c}{50.00 } & \multicolumn{1}{c}{\cellcolor[gray]{0.9}43.72 } \\
				\cmidrule{2-6}      & CViT \cite{CViT}  & \multicolumn{1}{c}{100} & \multicolumn{1}{c}{\cellcolor[gray]{0.9}100} & \multicolumn{1}{c}{50.00 } & \multicolumn{1}{c}{\cellcolor[gray]{0.9}60.20 } \\
				& EViT \cite{CEViT}  & \multicolumn{1}{c}{99.89 } & \multicolumn{1}{c}{\cellcolor[gray]{0.9}100} & \multicolumn{1}{c}{50.02 } & \multicolumn{1}{c}{\cellcolor[gray]{0.9}61.66 } \\
				& CEViT \cite{CEViT} & \multicolumn{1}{c}{99.98 } & \multicolumn{1}{c}{\cellcolor[gray]{0.9}100} & \multicolumn{1}{c}{50.50 } & \multicolumn{1}{c}{\cellcolor[gray]{0.9}86.45 } \\
				& SFDG \cite{SFDG} &\multicolumn{1}{c}{99.95 } & \multicolumn{1}{c}{\cellcolor[gray]{0.9}100 } & \multicolumn{1}{c}{50.08} & \multicolumn{1}{c}{\cellcolor[gray]{0.9}62.74} \\
				& GFF \cite{gff} &\multicolumn{1}{c}{99.98 } & \multicolumn{1}{c}{\cellcolor[gray]{0.9}100 } & \multicolumn{1}{c}{50.15} & \multicolumn{1}{c}{\cellcolor[gray]{0.9}63.21} \\
				& GramNet \cite{gramnet} &\multicolumn{1}{c}{99.95 } & \multicolumn{1}{c}{\cellcolor[gray]{0.9}100 } & \multicolumn{1}{c}{50.23} & \multicolumn{1}{c}{\cellcolor[gray]{0.9}70.15} \\
				
					& DIRE \cite{DIRE} &\multicolumn{1}{c}{99.99 } & \multicolumn{1}{c}{\cellcolor[gray]{0.9}100 } & \multicolumn{1}{c}{50.57} & \multicolumn{1}{c}{\cellcolor[gray]{0.9}87.12} \\	
					
				&  UFD \cite{UniversalFakeDetect} &\multicolumn{1}{c}{99.95 } & \multicolumn{1}{c}{\cellcolor[gray]{0.9}100 } & \multicolumn{1}{c}{50.34} & \multicolumn{1}{c}{\cellcolor[gray]{0.9}87.35} \\	
				
				& \textbf{Ours} & \multicolumn{1}{c}{\textbf{100 }} & \multicolumn{1}{c}{\textbf{\cellcolor[gray]{0.9}100 }} & \multicolumn{1}{c}{\textbf{50.60 }} & \multicolumn{1}{c}{\textbf{\cellcolor[gray]{0.9}92.73 }} \\
				\midrule
				\multirow{10}[6]{*}{GAN-syn.} & ResNet \cite{resnet} & \multicolumn{1}{c}{49.99 } & \multicolumn{1}{c}{\cellcolor[gray]{0.9}54.45 } & \multicolumn{1}{c}{99.79 } & \multicolumn{1}{c}{\cellcolor[gray]{0.9}99.99 } \\
				& Xception \cite{Xception} & \multicolumn{1}{c}{49.99 } & \multicolumn{1}{c}{\cellcolor[gray]{0.9}54.45 } & \multicolumn{1}{c}{99.79 } & \multicolumn{1}{c}{\cellcolor[gray]{0.9}99.99 } \\
				& VGG \cite{vgg}    & \multicolumn{1}{c}{49.81 } & \multicolumn{1}{c}{\cellcolor[gray]{0.9}61.61 } & \multicolumn{1}{c}{99.66 } & \multicolumn{1}{c}{\cellcolor[gray]{0.9}99.99 } \\
				\cmidrule{2-6}      & ViT \cite{ViT}   & \multicolumn{1}{c}{49.62 } & \multicolumn{1}{c}{\cellcolor[gray]{0.9}39.06 } & \multicolumn{1}{c}{98.67 } & \multicolumn{1}{c}{\cellcolor[gray]{0.9}99.92 } \\
				& CrossViT \cite{CrossViT} & \multicolumn{1}{c}{49.92 } & \multicolumn{1}{c}{\cellcolor[gray]{0.9}44.29 } & \multicolumn{1}{c}{98.69 } & \multicolumn{1}{c}{\cellcolor[gray]{0.9}99.88 } \\
				& iFormer \cite{Inceptiontransformer} & \multicolumn{1}{c}{49.82} & \multicolumn{1}{c}{\cellcolor[gray]{0.9}45.21} & \multicolumn{1}{c}{99.82} & \multicolumn{1}{c}{\cellcolor[gray]{0.9}100} \\
				\cmidrule{2-6}      & CViT \cite{CViT}   & \multicolumn{1}{c}{50.03 } & \multicolumn{1}{c}{\cellcolor[gray]{0.9}50.29 } & \multicolumn{1}{c}{99.81 } & \multicolumn{1}{c}{\cellcolor[gray]{0.9}99.98 } \\
				& EViT \cite{CEViT}  & \multicolumn{1}{c}{49.87 } & \multicolumn{1}{c}{\cellcolor[gray]{0.9}43.94 } & \multicolumn{1}{c}{100} & \multicolumn{1}{c}{\cellcolor[gray]{0.9}100} \\
				& CEViT \cite{CEViT} &\multicolumn{1}{c}{50.09 } & \multicolumn{1}{c}{\cellcolor[gray]{0.9}75.89 } & \multicolumn{1}{c}{100} & \multicolumn{1}{c}{\cellcolor[gray]{0.9}99.88} \\
				& SFDG \cite{SFDG}  &\multicolumn{1}{c}{50.07 } & \multicolumn{1}{c}{\cellcolor[gray]{0.9}52.35 } & \multicolumn{1}{c}{100} & \multicolumn{1}{c}{\cellcolor[gray]{0.9}99.96} \\
				& GFF \cite{gff}&\multicolumn{1}{c}{50.09 } & \multicolumn{1}{c}{\cellcolor[gray]{0.9}53.12 } & \multicolumn{1}{c}{100} & \multicolumn{1}{c}{\cellcolor[gray]{0.9}99.98} \\
				& GramNet \cite{gramnet} &\multicolumn{1}{c}{50.12 } & \multicolumn{1}{c}{\cellcolor[gray]{0.9}75.29 } & \multicolumn{1}{c}{100} & \multicolumn{1}{c}{\cellcolor[gray]{0.9}100} \\	
					& DIRE \cite{DIRE} &\multicolumn{1}{c}{50.08} & \multicolumn{1}{c}{\cellcolor[gray]{0.9}74.78} & \multicolumn{1}{c}{100} & \multicolumn{1}{c}{\cellcolor[gray]{0.9}100} \\	
						&  UFD \cite{UniversalFakeDetect} &\multicolumn{1}{c}{50.10 } & \multicolumn{1}{c}{\cellcolor[gray]{0.9}75.68 } & \multicolumn{1}{c}{100} & \multicolumn{1}{c}{\cellcolor[gray]{0.9}99.97} \\

				& \textbf{Ours}&\multicolumn{1}{c}{\textbf{50.13} } &
				 \multicolumn{1}{c}{\textbf{\cellcolor[gray]{0.9}83.02} } & \multicolumn{1}{c}{\textbf{100}} & \multicolumn{1}{c}{\textbf{\cellcolor[gray]{0.9}100}} \\
				\bottomrule     
			\end{tabular}%
		}

		\label{tab38}
	}
\end{table}%

\section{Experiments}\label{sec5}
\subsection{ Implementation Details}\label{subsec51}

We developed the detector using PyTorch on the Tesla V100 GPU with batch size 32. The number of blocks $K$, $S$, $L$, $E$, and $N$ in CAEL is set to 4, 2, 3, 3, and 2, respectively.
The attention heads $m$ are set to 8 and the dimension of the feature token $d$ is set to 192. We employed the Albumentations \cite{Albumentations} library to augment our data for training. We adopted a series of transformations including the random rotation, transpose, horizontal flip, vertical flip, and shift scale rotation, with a 0.9 probability. Our model is trained with the Adam optimizer \cite{Adam} with a learning rate of 1e-4 and weight decay of 1e-4. We utilized the scheduler to drop the learning rate by ten times every 15 epochs. We leveraged the cross-entropy loss function to train our model. 

{\bfseries\setlength\parindent{0em}  Evaluation Metrics.} We employed Accuracy (ACC) and Area Under the Receiver Operating Characteristic Curve (AUC) as our evaluation metrics. We further utilized precision, recall, and F1-score to evaluate the model performance. We computed the number of parameters and Floating Point Operations per Second (FLOPs) to assess the detection efficiency of different models.

\begin{table}[t]
	\centering
		\caption{Cross-generator evaluation. ACC and AUC scores (\%) on each facial attribute edited by IAFaceS, after training using images manipulated by Latenttransformer. 
	}	
	\setlength{\tabcolsep}{0.4mm}{
		\resizebox{0.48\textwidth}{!}{%
			\normalsize
			\begin{tabular}{cccccccccccc}
				
				\toprule
				\multirow{3}[6]{*}{Model} & \multicolumn{10}{c}{Attribute Manipulation}                                   &  \\
				\cmidrule{2-12}      & \multicolumn{2}{c}{bags} & \multicolumn{2}{c}{beard} & \multicolumn{2}{c}{bushy} & \multicolumn{2}{c}{\shortstack{\vspace{-0.40em}open\\ mouth} } & \multicolumn{2}{c}{\shortstack{\vspace{-0.05em} narrow\\ eyes}} &\multirow{2}[2]{*}{\shortstack{Average \\ AUC}} \\
				\cmidrule{2-11}      & ACC   & \cellcolor[gray]{0.9}AUC   & ACC   & \cellcolor[gray]{0.9}AUC   & ACC   & \cellcolor[gray]{0.9}AUC   & ACC   & \cellcolor[gray]{0.9}AUC   & ACC   & \cellcolor[gray]{0.9}AUC   &  \\
				\midrule
				ResNet \cite{resnet} & 50.00  & \cellcolor[gray]{0.9}49.31  & 50.00  &\cellcolor[gray]{0.9} 60.30  & 50.00  & \cellcolor[gray]{0.9}50.07  & 50.00  & \cellcolor[gray]{0.9}52.79  & 50.00  &\cellcolor[gray]{0.9} 53.24  & 53.14  \\
				Xception \cite{Xception} & 49.95  & \cellcolor[gray]{0.9}47.88  & \textbf{50.10} & \cellcolor[gray]{0.9}40.13  & 50.00  & \cellcolor[gray]{0.9}52.15  & 49.95  & \cellcolor[gray]{0.9}49.64  & 50.00  & \cellcolor[gray]{0.9}52.22  & 48.40  \\
				VGG \cite{vgg}   & 49.95  & \cellcolor[gray]{0.9}48.18  & 49.95  & \cellcolor[gray]{0.9}46.22  & 49.95  & \cellcolor[gray]{0.9}49.88  & 49.95  & \cellcolor[gray]{0.9}50.62  & 50.00  & \cellcolor[gray]{0.9}64.69  &51.92  \\
				ViT \cite{ViT}   & 49.90  & \cellcolor[gray]{0.9}44.93  & 49.85  & \cellcolor[gray]{0.9}45.40  & 49.80  & \cellcolor[gray]{0.9}50.72  & 49.85  & \cellcolor[gray]{0.9}51.12  & \textbf{50.50} & \cellcolor[gray]{0.9}47.60  & 47.95  \\
				CrossViT \cite{CrossViT} & \textbf{50.05} & \cellcolor[gray]{0.9}46.93  & 50.00  & \cellcolor[gray]{0.9}48.75  & 49.95  & \cellcolor[gray]{0.9}50.77  & 49.85  & \cellcolor[gray]{0.9}49.36  & 50.10  & \cellcolor[gray]{0.9}45.58  &48.28  \\
				iFormer \cite{Inceptiontransformer} & 50.00  & \cellcolor[gray]{0.9}54.14  & 50.00  & \cellcolor[gray]{0.9}53.28  & 50.00  &\cellcolor[gray]{0.9}53.89  & 50.00  & \cellcolor[gray]{0.9}55.70  & 50.00  &\cellcolor[gray]{0.9}56.20  &54.64  \\
				CViT \cite{CViT}  & \textbf{50.05} & \cellcolor[gray]{0.9}50.88  & 50.00  & \cellcolor[gray]{0.9}44.63  & 50.05  &\cellcolor[gray]{0.9}50.37  & 50.00  &\cellcolor[gray]{0.9} 51.27  & 50.00  & \cellcolor[gray]{0.9}49.58  &49.35  \\
				EViT \cite{CEViT}  & 50.00  & \cellcolor[gray]{0.9}53.62  & 50.05  & \cellcolor[gray]{0.9}51.55  & 50.00  & \cellcolor[gray]{0.9}55.19  & 50.00  & \cellcolor[gray]{0.9}54.10  & 50.00  & \cellcolor[gray]{0.9}54.96  &53.88  \\
				CEViT \cite{CEViT} & 50.00  & \cellcolor[gray]{0.9}55.57  & 50.00  & \cellcolor[gray]{0.9}49.31  & 50.00  & \cellcolor[gray]{0.9}57.61  & \textbf{50.05} & \cellcolor[gray]{0.9}56.23  & 50.00  & \cellcolor[gray]{0.9}55.32  & 54.81  \\
				SFDG \cite{SFDG}  & 50.00  & \cellcolor[gray]{0.9}54.65  & 50.01  & \cellcolor[gray]{0.9}50.54 & 50.00  & \cellcolor[gray]{0.9}55.61  & 50.03 & \cellcolor[gray]{0.9}54.23  & 50.09  & \cellcolor[gray]{0.9}54.32  & 55.87  \\
				GFF \cite{gff} & 50.00  & \cellcolor[gray]{0.9}60.14  & 50.00 & \cellcolor[gray]{0.9}50.65  & 50.02  & \cellcolor[gray]{0.9}57.14  & 50.02 & \cellcolor[gray]{0.9}56.17  & 50.03  & \cellcolor[gray]{0.9}55.62  & 57.81  \\
				GramNet \cite{gramnet} & 50.01  & \cellcolor[gray]{0.9}60.73  & 50.02  & \cellcolor[gray]{0.9}59.31  & 50.05  & \cellcolor[gray]{0.9}57.52  & 50.04 & \cellcolor[gray]{0.9}59.03  & 50.13  & \cellcolor[gray]{0.9}60.21  & 63.65  \\
				
					DIRE \cite{DIRE} & 50.04 & \cellcolor[gray]{0.9}61.07  & 50.03  & \cellcolor[gray]{0.9}60.15  & 50.04  & \cellcolor[gray]{0.9}59.41  & 50.03 & \cellcolor[gray]{0.9}60.74  & 50.34  & \cellcolor[gray]{0.9}61.32  & 63.47  \\
						UFD \cite{UniversalFakeDetect}  & 50.02  & \cellcolor[gray]{0.9}60.41  & 50.01  & \cellcolor[gray]{0.9}60.36  & 50.04  & \cellcolor[gray]{0.9}60.41  & 50.03& \cellcolor[gray]{0.9}60.21  & 50.15  & \cellcolor[gray]{0.9}60.25 & 64.36  \\

				\textbf{Ours} & 50.00  & \cellcolor[gray]{0.9}\textbf{69.84} & 50.00  & \cellcolor[gray]{0.9}\textbf{77.35} & \textbf{50.10} & \cellcolor[gray]{0.9}\textbf{73.43} & \textbf{50.05} &\cellcolor[gray]{0.9}\textbf{71.09} & 50.00  & \cellcolor[gray]{0.9}\textbf{71.62} & \textbf{72.67} \\
			
				\bottomrule
			\end{tabular}%
		}
	
		\label{tab4}
	}
\end{table}%

\begin{table}[t]
	\centering
	\caption{Cross-forgery generalization. ACC and AUC scores (\%) on each manipulation, after training using one manipulation. 
	}
	\setlength{\tabcolsep}{1.5mm}{
		\resizebox{0.48\textwidth}{!}{%
			\normalsize
			\begin{tabular}{cccccccc}
				\toprule
				\multirow{3}[6]{*}{\shortstack{Training\\ Set}} & \multirow{3}[6]{*}{Model} & \multicolumn{6}{c}{Testing Set } \\
				\cmidrule{3-8}      &       & \multicolumn{2}{c}{EFS} & \multicolumn{2}{c}{AM} & \multicolumn{2}{c}{FS} \\
				\cmidrule{3-8}      &       & ACC   &  \cellcolor[gray]{0.9}AUC   & ACC   &  \cellcolor[gray]{0.9}AUC   & ACC   &  \cellcolor[gray]{0.9}AUC \\
				\midrule
				\multirow{15}[6]{*}{EFS} & ResNet \cite{resnet} & 100   &  \cellcolor[gray]{0.9}100   & 50.00  &  \cellcolor[gray]{0.9}46.86 & 74.98  &  \cellcolor[gray]{0.9}70.75 \\
				& Xception \cite{Xception} & 100   &  \cellcolor[gray]{0.9}100   & 50.00  &  \cellcolor[gray]{0.9}63.14 & 68.06  &  \cellcolor[gray]{0.9}79.52 \\
				& VGG \cite{vgg}   & 100   &  \cellcolor[gray]{0.9}100   & 50.00  &  \cellcolor[gray]{0.9}67.02 & 75.00  &  \cellcolor[gray]{0.9}75.27  \\
				\cmidrule{2-8}      & ViT \cite{ViT}   & 95.31 &  \cellcolor[gray]{0.9}99.81 & \textbf{ 54.69 } &  \cellcolor[gray]{0.9}65.86  & 53.13  &  \cellcolor[gray]{0.9}61.43  \\
				& CrossViT \cite{CrossViT} & 98.46 &  \cellcolor[gray]{0.9}99.93 & 49.89  & \cellcolor[gray]{0.9}53.65 & 69.23  & \cellcolor[gray]{0.9}74.53  \\
				& iFormer \cite{Inceptiontransformer} & 100  & \cellcolor[gray]{0.9}100 & 50.00  &  \cellcolor[gray]{0.9}62.66 & 74.96  & \cellcolor[gray]{0.9} 73.21  \\
				\cmidrule{2-8}      & CViT \cite{CViT}   & 99.98 & \cellcolor[gray]{0.9}100   & 50.02  &  \cellcolor[gray]{0.9}63.53 & 72.79  &  \cellcolor[gray]{0.9}73.82  \\
				& EViT \cite{CEViT}  & 99.36 & \cellcolor[gray]{0.9}100   & 50.03  &  \cellcolor[gray]{0.9}70.58 & 74.61  &  \cellcolor[gray]{0.9}78.58  \\
				& CEViT \cite{CEViT} & 99.99 &  \cellcolor[gray]{0.9}100   & 50.00  &  \cellcolor[gray]{0.9}75.41 & 74.95  &  \cellcolor[gray]{0.9}78.32  \\
				& SFDG \cite{SFDG}  & 99.64  & \cellcolor[gray]{0.9}100  & 50.02 & \cellcolor[gray]{0.9}73.54 & 73.82  &\cellcolor[gray]{0.9}75.93 \\
				& GFF \cite{gff} & 99.52  & \cellcolor[gray]{0.9}100  & 50.01 &\cellcolor[gray]{0.9}74.32 & 73.91  & \cellcolor[gray]{0.9}76.43\\
				& GramNet \cite{gramnet} & 99.61  & \cellcolor[gray]{0.9}100  & 50.05 &\cellcolor[gray]{0.9} 76.24 & 74.21  & \cellcolor[gray]{0.9}76.13 \\
				& DIRE \cite{DIRE} & 99.94  & \cellcolor[gray]{0.9}100  & 50.03 &\cellcolor[gray]{0.9} 76.14 & 74.03  & \cellcolor[gray]{0.9}77.89 \\
				& UFD \cite{UniversalFakeDetect} & 99.98  & \cellcolor[gray]{0.9}100  & 50.04 &\cellcolor[gray]{0.9} 76.32 & 74.76 & \cellcolor[gray]{0.9}78.01 \\	
				& \textbf{Ours} & \textbf{100} &  \cellcolor[gray]{0.9}\textbf{100} & 50.04  &  \cellcolor[gray]{0.9}\textbf{ 83.31} & \textbf{ 75.00 } &  \cellcolor[gray]{0.9}\textbf{79.96} \\
				\midrule
				\multirow{15}[6]{*}{AM} & ResNet \cite{resnet} & 50.17  & \cellcolor[gray]{0.9}71.06  & 99.98  & \cellcolor[gray]{0.9}100   & 49.98  & \cellcolor[gray]{0.9}69.36  \\
				& Xception \cite{Xception} & 50.20  & \cellcolor[gray]{0.9}51.45  & 99.89  & \cellcolor[gray]{0.9}100   & 50.11  &\cellcolor[gray]{0.9} 54.57  \\
				& VGG \cite{vgg}  & 50.04  & \cellcolor[gray]{0.9}75.24  & 99.95  &\cellcolor[gray]{0.9}100   & 50.00  & \cellcolor[gray]{0.9}69.20  \\
				\cmidrule{2-8}      & ViT \cite{ViT}   & 50.29  & \cellcolor[gray]{0.9}60.37  & 99.63  & \cellcolor[gray]{0.9}99.99  & 50.19  &\cellcolor[gray]{0.9}55.04  \\
				& CrossViT \cite{CrossViT} & \textbf{51.21}  & \cellcolor[gray]{0.9}69.79  & 99.11  & \cellcolor[gray]{0.9}99.98  & \textbf{52.66}  &\cellcolor[gray]{0.9}77.68  \\
				& iFormer \cite{Inceptiontransformer} & 50.20  & \cellcolor[gray]{0.9}72.38  & 99.98  &\cellcolor[gray]{0.9}100   & 50.42  &\cellcolor[gray]{0.9}78.27  \\
				\cmidrule{2-8}      & CViT \cite{CViT}   & 50.15  & \cellcolor[gray]{0.9}70.32  & 99.89 & \cellcolor[gray]{0.9}100   & 50.02  &\cellcolor[gray]{0.9}60.74  \\
				& EViT \cite{CEViT}  &   50.26    & \cellcolor[gray]{0.9}56.65     &     99.91  &\cellcolor[gray]{0.9}100      &   49.95    &\cellcolor[gray]{0.9}57.28 \\
				& CEViT \cite{CEViT} & 50.13  & \cellcolor[gray]{0.9}73.38  & 99.98 & \cellcolor[gray]{0.9}100   & 50.07  & \cellcolor[gray]{0.9}80.06  \\
				& SFDG \cite{SFDG}  &50.15  & \cellcolor[gray]{0.9}60.73  & 99.93 & \cellcolor[gray]{0.9}100 & 50.03  &\cellcolor[gray]{0.9} 59.37 \\
				& GFF \cite{gff} & 51.02  & \cellcolor[gray]{0.9}62.75  & 99.96 &\cellcolor[gray]{0.9} 100 & 50.08  & \cellcolor[gray]{0.9}59.21 \\
				& GramNet \cite{gramnet} & 51.10  & \cellcolor[gray]{0.9}63.25  & 99.95 &\cellcolor[gray]{0.9} 100 & 50.09  & \cellcolor[gray]{0.9}60.35 \\
				& DIRE \cite{DIRE} & 51.14  & \cellcolor[gray]{0.9}72.41  & 99.94 &\cellcolor[gray]{0.9} 100 & 51.24  & \cellcolor[gray]{0.9}70.45 \\
				& UFD \cite{UniversalFakeDetect} & 50.65  & \cellcolor[gray]{0.9}73.40  & 98.65 &\cellcolor[gray]{0.9} 100 & 52.36  & \cellcolor[gray]{0.9}71.89 \\
				& \textbf{Ours} &50.10  & \cellcolor[gray]{0.9}\textbf{ 87.46} & \textbf{ 100} & \cellcolor[gray]{0.9}\textbf{ 100} & 50.19  & \cellcolor[gray]{0.9}\textbf{82.35} \\
				\midrule
				\multirow{15}[6]{*}{FS} & ResNet \cite{resnet}  & 50.24  & \cellcolor[gray]{0.9}73.13  & 50.41 & \cellcolor[gray]{0.9}76.85 & 99.88  & \cellcolor[gray]{0.9}100 \\
				& Xception \cite{Xception} & 50.42  & \cellcolor[gray]{0.9}76.48  & 53.75 & \cellcolor[gray]{0.9}75.62 & 99.84  &\cellcolor[gray]{0.9} 99.99 \\
				& VGG \cite{vgg}  & 50.22  &\cellcolor[gray]{0.9}72.37  & 49.98 &\cellcolor[gray]{0.9}79.66 & 99.98  &\cellcolor[gray]{0.9}100 \\
				\cmidrule{2-8}      & ViT \cite{ViT}   & 51.09  &\cellcolor[gray]{0.9}69.16  & 52.37 &\cellcolor[gray]{0.9}78.11 & 99.19  & \cellcolor[gray]{0.9}99.97 \\
				& CrossViT \cite{CrossViT} & 50.59  &\cellcolor[gray]{0.9}62.06  & 50.69 & \cellcolor[gray]{0.9}66.02 & 98.85  & \cellcolor[gray]{0.9}99.94 \\
				& iFormer \cite{Inceptiontransformer} &   50.63    &  \cellcolor[gray]{0.9}72.56    &  \textbf{53.84}    & \cellcolor[gray]{0.9}  \textbf{80.94}   &  99.93     &  \cellcolor[gray]{0.9}100\\
				\cmidrule{2-8}      & CViT \cite{CViT}  & 50.22  & \cellcolor[gray]{0.9}73.88  & 49.98 &\cellcolor[gray]{0.9} 73.75 & 99.96  &\cellcolor[gray]{0.9} 100 \\
				& EViT \cite{CEViT}  & 53.50  & \cellcolor[gray]{0.9}80.68  & 50.21 & \cellcolor[gray]{0.9}64.36 & 99.50  &\cellcolor[gray]{0.9} 99.99 \\
				& CEViT \cite{CEViT} & 53.64  & \cellcolor[gray]{0.9}80.96  & 50.45 &\cellcolor[gray]{0.9}73.84 & 99.97  & \cellcolor[gray]{0.9}100 \\
				& SFDG \cite{SFDG}  & 53.65  & \cellcolor[gray]{0.9}80.71  & 50.15 & \cellcolor[gray]{0.9}66.72 & 99.63 &\cellcolor[gray]{0.9} 100 \\
				& GFF \cite{gff} & 53.71  & \cellcolor[gray]{0.9}81.27  & 51.34 &\cellcolor[gray]{0.9} 68.25 & 99.76  & \cellcolor[gray]{0.9}100 \\
				& GramNet \cite{gramnet} & 54.83  & \cellcolor[gray]{0.9}80.89  & 51.19 &\cellcolor[gray]{0.9} 67.13 & 99.74  & \cellcolor[gray]{0.9}100 \\
				& DIRE \cite{DIRE} & 54.06  & \cellcolor[gray]{0.9}79.65  & 52.13 &\cellcolor[gray]{0.9} 78.32 & 99.54  & \cellcolor[gray]{0.9}99.98\\
				& UFD \cite{UniversalFakeDetect} & 53.48  & \cellcolor[gray]{0.9}80.62  & 53.45 &\cellcolor[gray]{0.9} 79.04 & 99.61  & \cellcolor[gray]{0.9}99.99 \\
				& \textbf{Ours} &  \textbf{56.65}  & \textbf{ \cellcolor[gray]{0.9} 88.41 } & 50.99 & \cellcolor[gray]{0.9} 78.32 &  \textbf{99.98} & \cellcolor[gray]{0.9}\textbf{100}\\
				\bottomrule
			\end{tabular}%
		}
		
		\label{tab6}
	}
	
\end{table}%

\subsection{GenFace Benchmark}\label{subsec52}
We conducted the within-dataset evaluation including cross-forgery evaluation and cross-generator evaluation, and the cross-dataset evaluation. 

We evaluated state-of-the-art deepfake detection approaches on our benchmark dataset. We selected common CNN-based models such as ResNet-18 \cite{resnet}, VGG-16 \cite{vgg}, Xception \cite{Xception}, general transformer-based approaches including ViT-B \cite{ViT}, CrossViT-B \cite{CrossViT}, and iFormer-B \cite{Inceptiontransformer} and some detectors such as UFD \cite{UniversalFakeDetect} and DIRE \cite{DIRE} which are specifically designed to detect diffusion-generated images. We also selected some hybrid transformer-based deepfake detectors such as CViT \cite{CViT}, EViT \cite{CEViT}, and CEViT \cite{CEViT} as well as the recent SOTA detectors including SFDG\cite{SFDG}, GFF\cite{gff}, and GramNet\cite{gramnet}.

{\bfseries\setlength\parindent{0em} Cross-forgery evaluation.} To evaluate our dataset, we conducted cross-forgery tests. We trained models using images of one manipulation, and tested them on those of each manipulation, i.e., EFS, AM, and FS, respectively. The results are displayed in Table~\ref{tab6}. We can observe that most models achieve poor performance (about 70\% AUC) on cross-forgery evaluation, which justifies the high image quality of our challenging GenFace dataset. For transformer-based detectors, the AUC of our network is around 21.6\%, 25.1\%, and 2.3\% higher than that of CViT, EViT, and CEViT, respectively, on FS after training using AM, which is attributed to the powerful fusion capabilities across multi-grained appearance and edge information.

{\bfseries\setlength\parindent{0em} Cross-generator evaluation.} To perform fine-grained analysis, we examined the performance of detectors on the cross-generator evaluation. We trained models using images of one generator from EFS, and tested them on different generators, i.e., diffusion-based and GAN-based generators, respectively. As Table~\ref{tab38} displays, models trained using diffusion-based generators outperform those trained with GAN-based methods. For instance, ResNet achieves 100\% AUC and 73.92\% AUC on the intra-generator and cross-generator testing, respectively, after training using images of the diffusion-based generator. By contrast, it only gains 99.99\% AUC and 54.45\% AUC when trained using the images of the GAN-based generator, which demonstrates that the diffusion-based generator tends to push models to study discriminative and comprehensive forgery artifacts, thus further enhancing detection accuracy. Models consistently achieve poorer performance on diffusion-based generators than those on GAN-based ones in Fig.~\ref{fig17}, on cross-generator evaluation, demonstrating that the diffusion-based generator produces higher-quality synthetic faces, making it more challenging for models to detect the forgery traces. Like \cite{UGID}, we observed that several results show around 50\% accuracy but a high AUC, showing that the usual decision threshold might not be trustful in the testing scenario. However, the predictions are still useful because they can mostly discern real from fake.

In order to further conduct the in-depth study of GenFace, we trained models using facial images with attribute manipulated by LatTrans, and tested them on five attributes manipulated by the IAFaceS generator, i.e., bags, beard, bushy, open mouth, and narrow eyes. In Table~\ref{tab4}, we noticed that the manipulated beard attribute is more difficult for detectors to identify, but the manipulation of two attributes such as narrow eyes and bushy are easier to detect. 

\begin{table}[t]
	\centering
	\caption{Impacts of various operators. MH means MarrHildreth. }	
	\setlength{\tabcolsep}{4.8mm}{
		\resizebox{0.44\textwidth}{!}{%
			\normalsize
			\begin{tabular}{cccc}
			\toprule
			\multirow{3}[6]{*}{\shortstack{Training\\ Set}} & \multirow{3}[6]{*}{Model} & \multicolumn{2}{c}{Testing Set } \\
			\cmidrule{3-4}      &       & \multicolumn{2}{c}{GAN-syn.} \\
			\cmidrule{3-4}      &       & ACC   & \cellcolor[gray]{0.9}AUC \\
			\midrule
			\multirow{5}[2]{*}{\shortstack{Diffusion\\-syn.}} & CAEL w/MH & 50.46  & \cellcolor[gray]{0.9}84.69  \\
			& CAEL  w/LoG & 50.05  & \cellcolor[gray]{0.9}48.59  \\
			& CAEL  w/DCT & 50.33  &\cellcolor[gray]{0.9} 90.21  \\
			& CAEL  w/Canny & 50.05  & \cellcolor[gray]{0.9}45.86  \\
			& CAEL  w/Sobel & \textbf{50.60} &\cellcolor[gray]{0.9} \textbf{92.73} \\
			\bottomrule
		\end{tabular}%
	}	
\label{tab5}
}
\end{table}

\begin{table}[t]
	\centering
		\caption{Network ablation. ACC and AUC scores (\%) of various settings on the GAN-based generator after training on the diffusion-based one.}
	\setlength{\tabcolsep}{9.2mm}{
		\resizebox{0.44\textwidth}{!}{%
			\normalsize
		\begin{tabular}{ccc}
	\toprule
	Model & ACC   & \cellcolor[gray]{0.9}AUC \\
	\midrule
	F     & 50.02  & \cellcolor[gray]{0.9}84.33  \\
	C     & 50.12  & \cellcolor[gray]{0.9}86.09  \\
	E     & 50.15  & \cellcolor[gray]{0.9}87.64  \\
	F+C   & 50.21  & \cellcolor[gray]{0.9}87.45  \\
	F+E   & 50.37  & \cellcolor[gray]{0.9}88.86  \\
	C+E   & 50.45  & \cellcolor[gray]{0.9}90.32  \\
	\textbf{F+C+E} & \textbf{ 50.60 } & \textbf{\cellcolor[gray]{0.9} 92.73 } \\
	\bottomrule
\end{tabular}%
}
		
		\label{tab:tab16}
	}
\end{table}

\begin{table*}[t]
	\centering
	\caption{Cross-dataset evaluation. ACC and AUC scores (\%) of different methods on GenFace, FF++, DFDC, Celeb-DF, and DF-1.0, after training on FF++.}
	\setlength{\tabcolsep}{1.5mm}{
		\resizebox{0.9\textwidth}{!}{%
			\normalsize
			\begin{tabular}{ccccccccccccccc}
				\toprule
				\multirow{3}[6]{*}{\shortstack{Training\\ Set}} & \multirow{3}[6]{*}{\hspace{-0.8em}Model} & \multicolumn{10}{c}{Testing Set} & \multicolumn{2}{c}{\multirow{2}[3]{*}{}} \\
				\cmidrule{3-14}      &       & \multicolumn{2}{c}{GenFace} & \multicolumn{2}{c}{FF++} & \multicolumn{2}{c}{DFDC} & \multicolumn{2}{c}{Celeb-DF} & \multicolumn{2}{c}{DF-1.0} & & \\
				\cmidrule{3-14}      &       & ACC   & \cellcolor[gray]{0.9}AUC   & ACC   & \cellcolor[gray]{0.9}AUC   & ACC   &\cellcolor[gray]{0.9} AUC   & ACC   &\cellcolor[gray]{0.9} AUC   & ACC   & \cellcolor[gray]{0.9}AUC & Params(M) & FLOPs(G) \\
				\midrule
				\multirow{9}[2]{*}{FF++} 
				& \hspace{-0.8em}Xception \cite{Xception} & 53.85 & \cellcolor[gray]{0.9}60.61 & 90.49 & \cellcolor[gray]{0.9}96.72 & 60.27 & \cellcolor[gray]{0.9}66.95 & 54.24 & \cellcolor[gray]{0.9}65.86 & 59.40 & \cellcolor[gray]{0.9}63.59 & 20.81 & 4.59 \\
				& \hspace{-0.8em}ResNet \cite{resnet} & 50.86 & \cellcolor[gray]{0.9}62.27 & 79.06 & \cellcolor[gray]{0.9}87.85 & \textbf{67.81} & \cellcolor[gray]{0.9}68.07 & 59.31 & \cellcolor[gray]{0.9}62.35 & 55.33 & \cellcolor[gray]{0.9}62.31 & 11.18 & 1.82 \\
				& \hspace{-0.5em}CrossViT \cite{CrossViT} & 51.10 & \cellcolor[gray]{0.9}58.21 & 62.49 & \cellcolor[gray]{0.9}67.50 & 54.05 & \cellcolor[gray]{0.9}58.46 & \textbf{63.50} & \cellcolor[gray]{0.9}59.72 & 49.55 & \cellcolor[gray]{0.9}59.30 & 104.70& 21.20 \\
				& \hspace{-0.8em}CViT \cite{CViT} & 56.99 & \cellcolor[gray]{0.9}61.96 & 90.25 & \cellcolor[gray]{0.9}96.27 & 51.04 & \cellcolor[gray]{0.9}63.98 & 60.61 & \cellcolor[gray]{0.9}64.96 & \textbf{62.15} & \cellcolor[gray]{0.9}64.53 & 89.02 & 6.69 \\
				& \hspace{-0.8em}SFDG \cite{SFDG} & 53.74 & \cellcolor[gray]{0.9}59.13 & 90.30 & \cellcolor[gray]{0.9}96.71 & 60.24 & \cellcolor[gray]{0.9}67.31 & 55.29 & \cellcolor[gray]{0.9}63.48 & 58.21 & \cellcolor[gray]{0.9}61.48 & 50.49 & 43.26 \\
				& \hspace{-0.5em}GFF \cite{gff} & 54.61 & \cellcolor[gray]{0.9}60.27 & 91.08 & \cellcolor[gray]{0.9}97.64 & 61.15 & \cellcolor[gray]{0.9}68.02 & 57.12 & \cellcolor[gray]{0.9}62.37 & 60.14 & \cellcolor[gray]{0.9}62.74 & 53.24 & 13.79 \\
				& \hspace{-0.5em}GramNet \cite{gramnet} & 54.72 & \cellcolor[gray]{0.9}61.38 & 90.45 & \cellcolor[gray]{0.9}96.85 & 60.37 & \cellcolor[gray]{0.9}68.14 & 56.24 & \cellcolor[gray]{0.9}62.67 & 59.38 & \cellcolor[gray]{0.9}63.82 & 11.69 & 1.82 \\
				& DIRE \cite{DIRE} & 55.74 & \cellcolor[gray]{0.9}61.25 & 90.86 & \cellcolor[gray]{0.9}97.85 & 62.45 & \cellcolor[gray]{0.9}66.74 & 59.75 & \cellcolor[gray]{0.9}60.12 & 60.41 & \cellcolor[gray]{0.9}63.87 & 23.51 & 4.13 \\
				& UFD \cite{UniversalFakeDetect} & 56.21 & \cellcolor[gray]{0.9}61.89 & 90.69 & \cellcolor[gray]{0.9}97.03 & 62.26 & \cellcolor[gray]{0.9}67.03 & 59.96 & \cellcolor[gray]{0.9}61.35 & 60.56 & \cellcolor[gray]{0.9}63.98 & 202.05 & 51.90 \\
				& \textbf{\hspace{-0.8em}Ours} & \textbf{58.55} & \cellcolor[gray]{0.9}\textbf{65.04} & \textbf{92.87} & \cellcolor[gray]{0.9}\textbf{98.11} & 64.64 & \cellcolor[gray]{0.9}\textbf{68.98} & 60.23 & \cellcolor[gray]{0.9}\textbf{66.48} & 58.85 & \cellcolor[gray]{0.9}\textbf{65.23} & 158.63 & 2.11 \\
				\bottomrule
			\end{tabular}%
		}
		\label{tab14}
	}
\end{table*}

{Cross-dataset evaluation.} To confirm the high quality of our dataset, we conducted the cross-dataset evaluation. We trained different deepfake detection approaches using the FF++ dataset and tested them on other datasets, i.e., GenFace, FF++, DFDC, Celeb-DF, and DF-1.0, respectively. The results are shown in Table~\ref{tab14}. We can see that models consistently show poorer performance on GenFace, compared to other datasets, demonstrating that our GenFace dataset is more challenging. 

Although the proposed CAEL model requires a larger number of parameters and FLOPs than lightweight ResNet18, the detection performance of CAEL is significantly improved on FF++ and GenFace. In particular, the ACC and AUC scores of CAEL are around 13.81\% and 10.26\% higher than that of ResNet18 on FF++, respectively, with increase of 147.45M parameters and 0.29G FLOPs. In the future work, we expect to apply further improvements, such as model architecture modifications and knowledge distillation, for faster efficiency. For the former, we can share weights across different layers of the transformer, to reduce the number of parameters and computational costs. For the latter, a smaller and faster student model can be trained to mimic the behavior of the proposed CAEL network.
 
\subsection{Ablation Study}
In order to investigate the contribution of each module to learning capacity, we observed the performance on cross-generator setting. As shown in Table~\ref{tab:tab16}, F, C, and E mean the fine-grained branch, coarse-grained branch, and edge branch, respectively. The gains from introducing the edge branch (+4.5\%) are evident, demonstrating that various edge representations provide valuable information to benefit deepfake detection. Especially, the AUC achieves the maximum after integrating multi-grained appearance embeddings with edge features.
\begin{figure}[t]%
	\includegraphics[width=\linewidth]{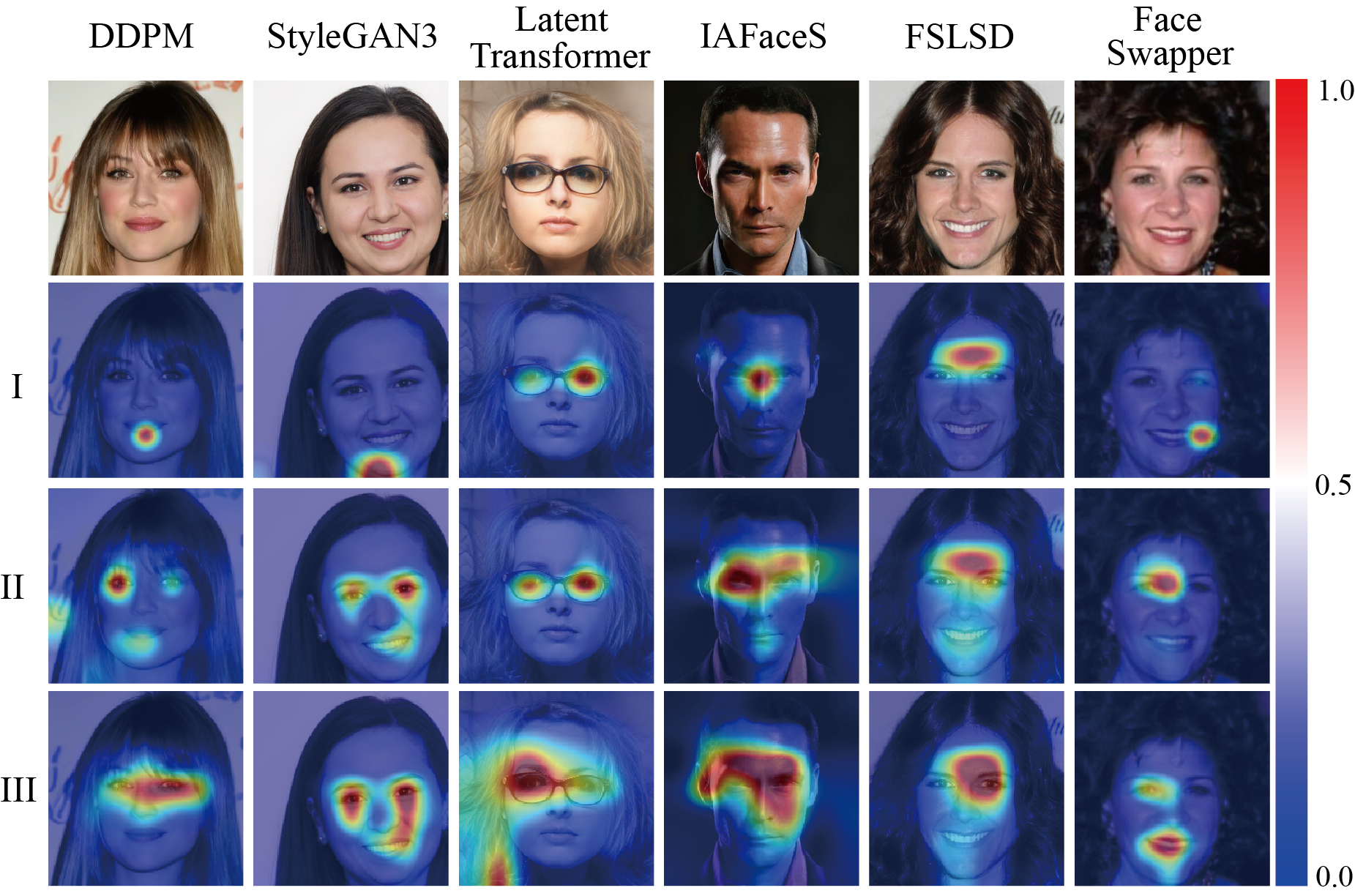}
	\centering
	\caption{The heatmap visualizations of various training settings on some examples from GenFace.}\label{fig9}
\end{figure}
\vspace{-0.2cm}
\begin{table}[t]
	\centering
	\caption{Case study. ACC scores (\%) of CAEL prediction and human evaluation on various face forgery images.}
	\setlength{\tabcolsep}{2.8mm}{
		\resizebox{0.48\textwidth}{!}{%
			\normalsize
\begin{tabular}{cccc}
	\toprule
	&       &\shortstack{ Model \\ Inference} & \shortstack{Human \\ Evaluation} \\
	\midrule
	Forgery Type & Forgery Method & ACC  & ACC \\
	\midrule
	\multirow{3}[2]{*}{FaceSwap} & DiffFace & 51.48 & 36.26 \\
	& FSLSD & 55.19 & 48.89 \\
	& FaceSwapper & 60.15 & 57.48 \\
	\midrule
	\multirow{3}[2]{*}{\shortstack{Attribute \\ Manipulation}  } & Diffae & 49.14 & 34.19 \\
	& IAFaceS & 50.03 & 45.67 \\
	& LatTrans& 54.21 & 50.39 \\
	\bottomrule
\end{tabular}%
 
		}
		
		\label{tab:tab26}
	}
\end{table}

\subsection{Case Study}
To build trust and transparency, we conducted a comprehensive case study to investigate the accuracy of both model inference and human evaluation on face images generated by different forgery approaches. We randomly selected images generated using various methods. To ensure a balanced and unbiased evaluation, we maintained an equal distribution of authentic and fake face images, i.e., 1:1 ratio, with each identity occurring merely once. We send 600 images to 80 participants and our CAEL model to make predictions.

The results across all manipulated methods under two forgery types are shown in Table~\ref{tab:tab26}. The proposed CAEL model consistently achieves superior ACC scores among all forgery approaches, compared to the human. In detail, we noticed that the accuracy of our CAEL model is significantly higher than that of human judgments on face images generated by DiffFace, mainly due to the reason that the diffusion model tends to yield photorealistic faces, which look real to human eyes.

\subsection{Visualization and Discussion}\label{subsec53}
We evaluated our detector by visualization, facial image frequency analysis on GenFace, robustness analysis. We investigated the effect of different operators, various queries, different fusion schemes, appearance-edge cross-attention, the depth of the edge transformer encoder, fine-grained classification, and the number of multi-grained appearance-edge transformer blocks, respectively.

{\bfseries\setlength\parindent{0em} Visualization.} To further investigate the effectiveness of our detector, we displayed the Gradient-weighted Class Activation Mapping (Grad-CAM) of the sample, generated by each generator for three networks in Fig.~\ref{fig9}. Each column shows a forgery face yielded by various generative methods, including DDPM, StyleGAN3, LatTrans, etc. The second to fourth rows illustrate heatmaps for three models listed in Table~\ref{tab6}: (I) the EViT model; (II) CEViT; and (III) our CAEL network. CEViT (II) detects more long-range forgery patterns than EViT (I), which demonstrates the effectiveness of the fusion across multi-grained appearance features. In comparison to (II), CAEL (III) further captures more manipulated areas, showing that comprehensively integrating the edge information with multi-grained appearance embeddings can benefit deepfake detection. Specifically, we added eyeglasses to pristine facial images through the attribute-manipulation model LatTrans, and we observed that the CAEL model can, to a large extent, identify the manipulated regions.

\begin{table}[t]
	\centering
		\caption{Effects of various queries.}	
	\setlength{\tabcolsep}{0.6mm}{
		\resizebox{0.44\textwidth}{!}{%
			\normalsize
			\begin{tabular}{cccccc}
				\toprule
				\multirow{3}[6]{*}{\shortstack{Training \\ Set}} & \multirow{3}[6]{*}{Model} & \multicolumn{4}{c}{Testing Set } \\
				\cmidrule{3-6}      &       & \multicolumn{4}{c}{GAN-syn.} \\
				\cmidrule{3-6}      &       & ACC \hspace{-0.1em} & \cellcolor[gray]{0.9}\hspace{0.1em} AUC  \hspace{0.4em}  & Params(M) & Flops(G) \\
				\midrule
				\multirow{3}[2]{*}{\shortstack{Diffusion\\-syn.}} & Query w/cls & \textbf{50.60} & \cellcolor[gray]{0.9}\textbf{92.73} & 158.63 & 2.11  \\
				& Query w/patch & 50.56  & \cellcolor[gray]{0.9}92.61  & 158.63  & 2.11  \\
				& Query w/all & 50.58  & \cellcolor[gray]{0.9}92.64  & 158.63  & 2.12  \\
				\bottomrule
			\end{tabular}%
		}

		\label{tab:tab19}	
	}

\end{table}

{\bfseries\setlength\parindent{0em} Impacts of different operators.}
We investigated the effect of various operators, including edge Sobel, Canny, LoG, and MarrHildreth, as well as the frequency-based DCT. We tested models on cross-generator setting. The results of ablation are shown in Table~\ref{tab5}. When Sobel edge is involved, the AUC achieves the maximum. We believe that edge features extracted by Sobel are discriminative, which can further improve the detection performance.

{\bfseries\setlength\parindent{0em} Influence of various queries.} For the AECA module of our method, we used the edge class token as the query. Next, we varied the query to see the effect on performance and efficiency. In Table~\ref{tab:tab19}, we noticed that the computational complexity of the model is increased by 0.01G, when all tokens are encompassed. Since the computational cost of creating an attention map is linear when the class token is used as a query, rather than quadratic like when all tokens serve as a query, this makes the whole process more efficient.
\begin{figure*}[t]
	\centering
	\includegraphics[width=\linewidth]{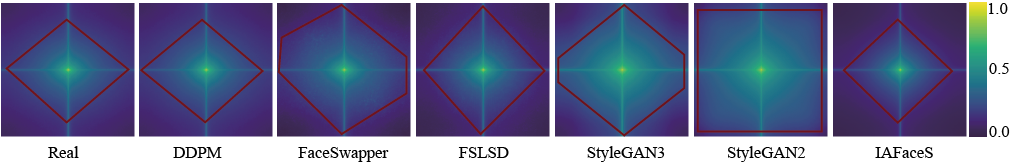}
	\centering
	\caption{The frequency visualization of facial images generated by various generators. The brighter the color, the greater the magnitude. }\label{figs2}
\end{figure*}

\begin{table}[t]
	\centering
	\caption{The classification performance at coarse-grained level.}	
	\setlength{\tabcolsep}{2.5mm}{
		\resizebox{0.45\textwidth}{!}{%
			\normalsize
			\begin{tabular}{cccccc}
				\toprule
				\multirow{2}[4]{*}{Model} & \multicolumn{5}{c}{Coarse-grained level} \\
				\cmidrule{2-6}      
				& ACC & AUC & Precision & Recall & F1 \\
				\midrule	
				ResNet \cite{resnet}  &  99.23 & 99.98 & 99.24 & 99.23 & 99.23 \\
				ViT \cite{ViT} & 96.32 & 96.94 & 96.50 & 96.51 & 96.51 \\
				CViT \cite{CViT} & 99.86 & 99.99 & 99.86 & 99.86 & 99.86 \\
				CEViT \cite{CEViT} & 99.87 & 99.99 & 99.88 & 99.88 & 99.87 \\
				\textbf{Ours} & \textbf{99.88} & \textbf{100} & \textbf{99.88} & \textbf{99.88} & \textbf{99.88} \\
				\bottomrule
			\end{tabular}%
		}
		\label{tab:tab39}	
	}
\end{table}

\begin{table}[t]
	\centering
		\caption{The classification performance at forgery level.}	
	\setlength{\tabcolsep}{3.2mm}{
		\resizebox{0.45\textwidth}{!}{%
			\normalsize
			\begin{tabular}{cccrr}
				\toprule
				\multirow{2}[4]{*}{Model} & \multicolumn{4}{c}{Forgery Level} \\
				\cmidrule{2-5}      & ACC   & Precision & \multicolumn{1}{c}{Recall} & \multicolumn{1}{c}{F1} \\
				\midrule	
				ResNet \cite{resnet}  &  99.78    &   99.78    &   99.77   & 99.77 \\
				Xception \cite{Xception} &   99.66   &   99.66    &    99.66   &  99.66 \\
				ViT\cite{ViT}  & 97.50     &   97.61   &    97.50   &  97.51 \\
				CViT \cite{CViT}  &   99.83   &   99.83    &    99.83   &  99.82 \\
				CEViT \cite{CEViT} &    99.88   &   99.88    &  99.88     & 99.87 \\
				\textbf{Ours} &    \textbf{99.95}   &  \textbf{ 99.95}    &   \textbf{99.95}    & \textbf{99.95} \\
				\bottomrule
			\end{tabular}%
		}
		\label{tab:tab30}	
	}
\end{table}

\begin{table}[t]
	\centering
		\caption{The classification performance at generator level.}
	\setlength{\tabcolsep}{3.2mm}{
		\resizebox{0.45\textwidth}{!}{%
			\normalsize
			\begin{tabular}{cccrr}
				\toprule
				\multirow{2}[3]{*}{Model} & \multicolumn{4}{c}{Generator Level} \\
				\cmidrule{2-5}      & ACC   & Precision & \multicolumn{1}{c}{Recall} & \multicolumn{1}{c}{F1} \\
				\midrule
				ResNet \cite{resnet}  &   99.67   &   99.67   &   99.67  & 99.67 \\
				Xception \cite{Xception}  &   99.84   &   99.84    &    99.84  &  99.84 \\
			   ViT\cite{ViT}  &  98.73  &  98.73  &  98.73    & 98.73 \\
				CViT \cite{CViT}  &  99.93    &    99.93   &  99.93     & 99.93 \\
				CEViT \cite{CEViT} &     99.94  &  99.94     &   99.94    &  99.94\\
				\textbf{Ours} &   \textbf{99.95}  &  \textbf{99.98}   &  \textbf{99.97}   & \textbf{99.97}\\
				\bottomrule
			\end{tabular}%
		}
		
		\label{tab:tab31}	
	}
\end{table}
\vspace{-1em}
{\bfseries\setlength\parindent{0em} Classification performance of various levels.} For coarse-grained detection, the experimental results are shown in Table~\ref{tab:tab39}. The forgery level and generator level detection results of various models are displayed in Table~\ref{tab:tab30} and Table~\ref{tab:tab31}, respectively. ACC scores indicate that the classification becomes easier with the number of categories growing from 2 (coarse-grained level) to 4 (forgery level) and 5 (generator level). Moreover, our model consistently shows excellent performance on various level classifications, compared to other baselines. Next, we plan to design a lightweight detector to jointly dig hierarchical clues.

\begin{table}[t]
	\centering
	\caption{The ablation of the AECA module.}	
	\label{tab:tab22}
	\setlength{\tabcolsep}{4mm}{
		\resizebox{0.47\textwidth}{!}{%
			\normalsize
			\begin{tabular}{cccc}
				\toprule
				\multirow{3}[6]{*}{Training Set} & \multirow{3}[6]{*}{Model} & \multicolumn{2}{c}{Testing Set } \\
				\cmidrule{3-4}      &       & \multicolumn{2}{c}{GAN-syn.} \\
				\cmidrule{3-4}      &       & ACC   &\cellcolor[gray]{0.9} AUC \\
				\midrule
				\multirow{2}[2]{*}{Diffusion-syn.} & ours w/o AECA & 50.47  & \cellcolor[gray]{0.9}91.35  \\
				& ours w/ AECA & \textbf{50.60}  & \cellcolor[gray]{0.9}\textbf{92.73}  \\
				\bottomrule
			\end{tabular}%
		}
		
	}
\end{table}

\begin{table}[t]
	\centering	
	\caption{The effect of the depth of the edge transformer encoder and the 	
		number of multi-grained appearance-edge transformer blocks. }		\label{tab21}
	\setlength{\tabcolsep}{1.9mm}{
		\resizebox{0.47\textwidth}{!}{%
			\normalsize
			\begin{tabular}{cccccc}
				\toprule
				\multirow{3}[6]{*}{\shortstack{Training \\ Set}} & \multirow{3}[6]{*}{Model} & \multicolumn{4}{c}{Testing Set } \\
				\cmidrule{3-6}      &       & \multicolumn{4}{c}{GAN-syn.} \\
				\cmidrule{3-6}      &       & \hspace{-0.2em}ACC   & \cellcolor[gray]{0.9}AUC \hspace{0.2em}  & \multicolumn{1}{l}{Params(M)} & Flops(G) \\
				\midrule
				\multirow{10}[4]{*}{\shortstack{Diffusion\\-syn.}} & E=0   & 50.12  & \cellcolor[gray]{0.9}87.33  & 130.27  & 2.05 \\
				& E=1  & 50.24  & \cellcolor[gray]{0.9}90.89  & 139.73  & 2.07 \\
				& E=2   & 50.49  & \cellcolor[gray]{0.9}91.78  & 149.18  & 2.09 \\
				& E=3   & \textbf{50.60} & \cellcolor[gray]{0.9}\textbf{92.73} & 158.63  & 2.11 \\
				& E=4   & 50.51  & \cellcolor[gray]{0.9}92.04  & 168.09 & 2.13 \\
				\cmidrule{2-6}      & K=1   & 50.09  & \cellcolor[gray]{0.9}86.75  & 92.14  & 1.99 \\
				& K=2   & 50.11  &\cellcolor[gray]{0.9}88.20  & 114.31  & 2.03 \\
				& K=3   & 50.44  &\cellcolor[gray]{0.9}90.65  & 136.47  & 2.07 \\
				& K=4   & \textbf{50.60} & \cellcolor[gray]{0.9}\textbf{92.73} & 158.63  & 2.11 \\
				& K=5   & 50.57  & \cellcolor[gray]{0.9}92.56  & 180.80 & 2.15 \\
				\bottomrule
			\end{tabular}%
		}
		
	}
\end{table}%

{\bfseries\setlength\parindent{0em} Effects of appearance-edge cross-attention.}
We further studied the importance of AECA through cross-generator evaluation. In Table~\ref{tab:tab22}, we noticed that a nearly 1.4\% increase of AUC could be achieved by introducing AECA. It turns out that the diverse global fusion between multi-grained appearance and edge domains is helpful for detecting comprehensive counterfeit traces.

{\bfseries\setlength\parindent{0em} Facial image frequency analysis.} We visualized the average spectrum of authentic and manipulated images from various generators. We utilized the Fourier transform to examine the frequency of both authentic and manipulated facial images. We randomly selected 5,000 images for Fourier transform, and averaged the results for spectral analysis. Fig.~\ref{figs2} illustrates the power spectrum of real and synthetic images, where the frequency increases gradually from the center to the edges, and the magnitude decreases, creating a smooth, radial gradient effect. We illustrated line profiles to compare the fall-off of spatial frequencies across different spectra. We can observe that the main difference in spectra is the rate at which the magnitude decreases with increasing spatial frequency. Besides, it is noticed that the spectrum of faces synthesized by the diffusion-based model is closer to that of the authentic face than GAN-based methods. Consequently, diffusion-based generators pose a greater threat to deepfake detection.
\begin{figure*}[t!]%
	\includegraphics[width=\linewidth]{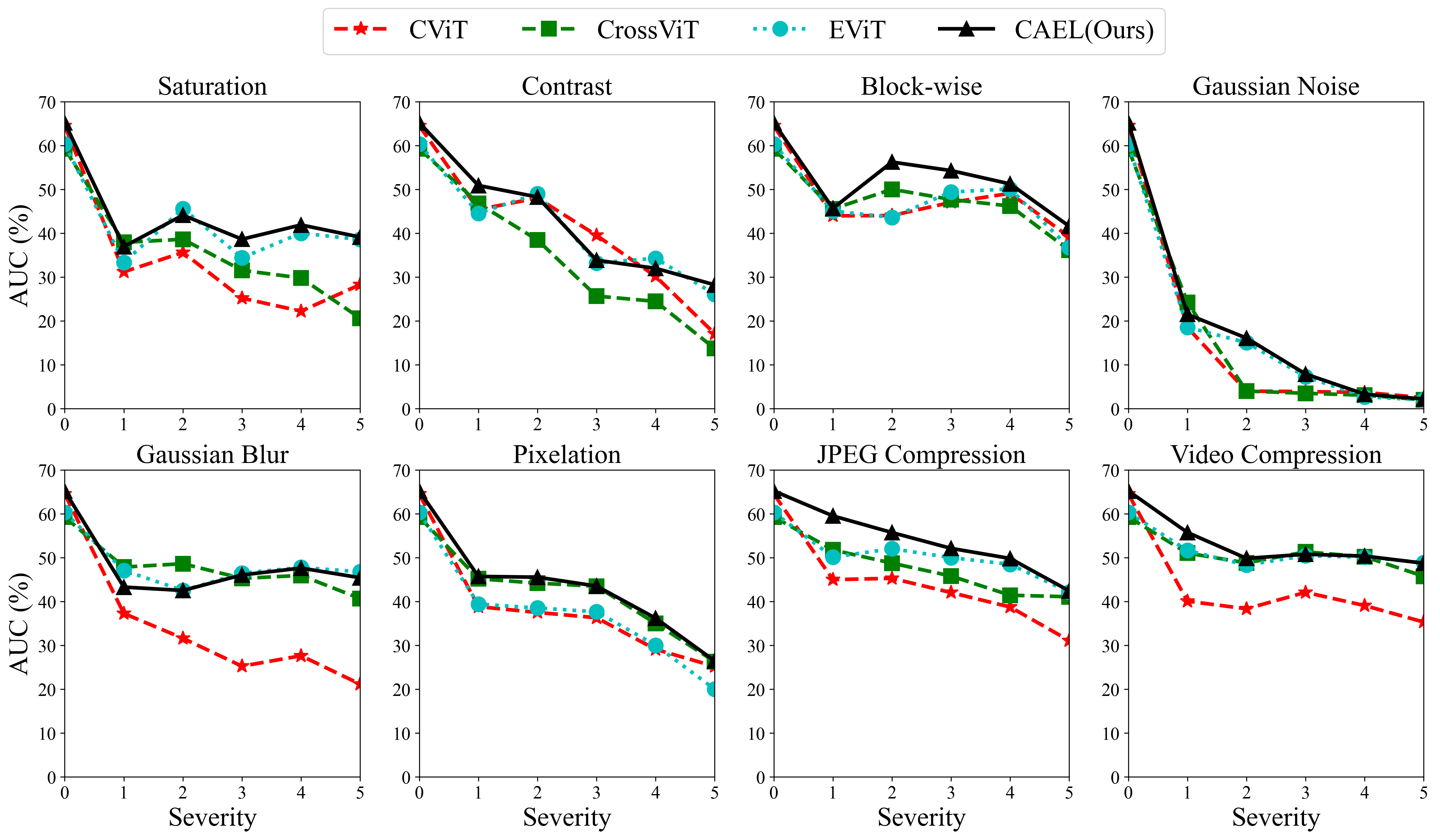}
	\centering
	\caption{Robustness to unseen image distortions.}\label{fig15}
\end{figure*}

{\bfseries\setlength\parindent{0em} Depth of edge transformer encoder and the number of multi-grained appearance-edge transformer blocks.} We inspected the effect of the depth of the edge transformer encoder. As shown in Table~\ref{tab21}, we observed that the performance of CAEL gradually grows with the increase of edge transformer blocks. The AUC achieves the maximum when three blocks are involved. We also observed the impacts of the number of the MAET on detection performance by varying the depth $K$, and other parameters remain unchanged. Using more than four transformer blocks in MAET increases FLOPs and parameters without any improvement in accuracy. We suspected that too many blocks would lead to information redundancy, thus reducing the representation ability.
\begin{table}[t]
	\centering
	\caption{The ablation of various fusions of different architectures.}	
	\label{tab:tab29}
	\setlength{\tabcolsep}{5mm}{
		\resizebox{0.45\textwidth}{!}{%
			\normalsize
			\begin{tabular}{cccc}
				\toprule
				Model & Fusion & ACC   &  \cellcolor[gray]{0.9}AUC \\
				\midrule
				\multirow{3}[2]{*}{F+E} & Concatenation &  50.15     & \cellcolor[gray]{0.9}86.61 \\
				& Summation &   50.26    & \cellcolor[gray]{0.9}87.72 \\
				& Cross-Attention &  50.37     & \cellcolor[gray]{0.9}88.86 \\
				\midrule
				\multirow{3}[2]{*}{C+E} & Concatenation &   50.29    &\cellcolor[gray]{0.9}88.17  \\
				& Summation &  50.37     &\cellcolor[gray]{0.9}89.21  \\
				& Cross-Attention &   50.45    & \cellcolor[gray]{0.9}90.32 \\
				\midrule
				\multirow{3}[2]{*}{F+C+E} & Concatenation &  50.48     & \cellcolor[gray]{0.9}90.95 \\
				& Summation &   50.54    & \cellcolor[gray]{0.9}91.56 \\
				& Cross-Attention & \textbf{50.60}      &\cellcolor[gray]{0.9}\textbf{92.73}  \\
				\bottomrule
			\end{tabular}%
		}
		
	}	
\end{table}

{\bfseries\setlength\parindent{0em} Robustness analysis.} We aimed to evaluate the robustness of detectors against diverse unseen image distortions by training them on GenFace, and testing their performance on various distorted images from \cite{DF1.0}. Eight types of perturbations are included, each of which is divided into five intensities. As Fig.~\ref{fig15} illustrates, we tested models on different image corruptions with a series of severities, including saturation adjustments, contrast modifications, block-wise distortions, white Gaussian noise addition, blurring, pixelation, JPEG compression, and video compression using the H.264 codec. A severity of 0 means that no degradation is applied.

Specifically, saturation adjustments aim to convert the image from BGR to YCbCr color space, adjust the chrominance components (Cb and Cr) by parameters to change the saturation, and then convert the image back to BGR color space. The parameters [0.4, 0.3, 0.2, 0.1, 0.0] correspond to five degrees of distortion (the smaller the parameter value, the greater the intensity). Contrast modifications intend to enlarge or reduce the image pixel value through parameters [0.85, 0.725, 0.6, 0.475, 0.35], which correspond to five intensities (the smaller the parameter value, the greater the distortion degree). Block-wise distortions draw fixed-size blocks at random positions on the image, and the number of blocks is controlled by the factor [16, 32, 48, 64, 80]. Gaussian noise addition intends to introduce different intensities of the Gaussian noise to the image via parameters [0.001, 0.002, 0.005, 0.01, 0.05]. Gaussian blur adjusts the standard deviation of the Gaussian kernel through the factors [7, 9, 13, 17, 21] (the larger the value, the more obvious the blurring effect). Pixelation reduces the image resolution by dividing the original image size by parameters [2, 3, 4, 5, 6] and then enlarges it back to the initial image size. JPEG compression is achieved by adjusting the compression quality parameters [20, 25, 30, 45, 40] (The larger the parameter value, the greater the compression rate). Similarly, video compression controls the compression quality of the video via the constant rate factor [30, 32, 35, 38, 40] (The larger the parameter value, the greater the compression rate). When applying corruptions of varying intensities, the changes of AUC for all techniques are shown in Fig.~\ref{fig15}. The statistics show that our model consistently outperforms all other methods for all types of image corruption. 

{\bfseries\setlength\parindent{0em} Comparison of different fusion schemes.} We further investigated the impact of various fusion schemes. Table~\ref{tab:tab29} shows the performance of various fusion strategies for different models. Among all the compared schemes, the proposed cross-attention strategies achieve the best AUC, showing the effectiveness of the diverse global fusion between appearance and edge features.

\begin{figure}[htbp]
	\includegraphics[width=\linewidth]{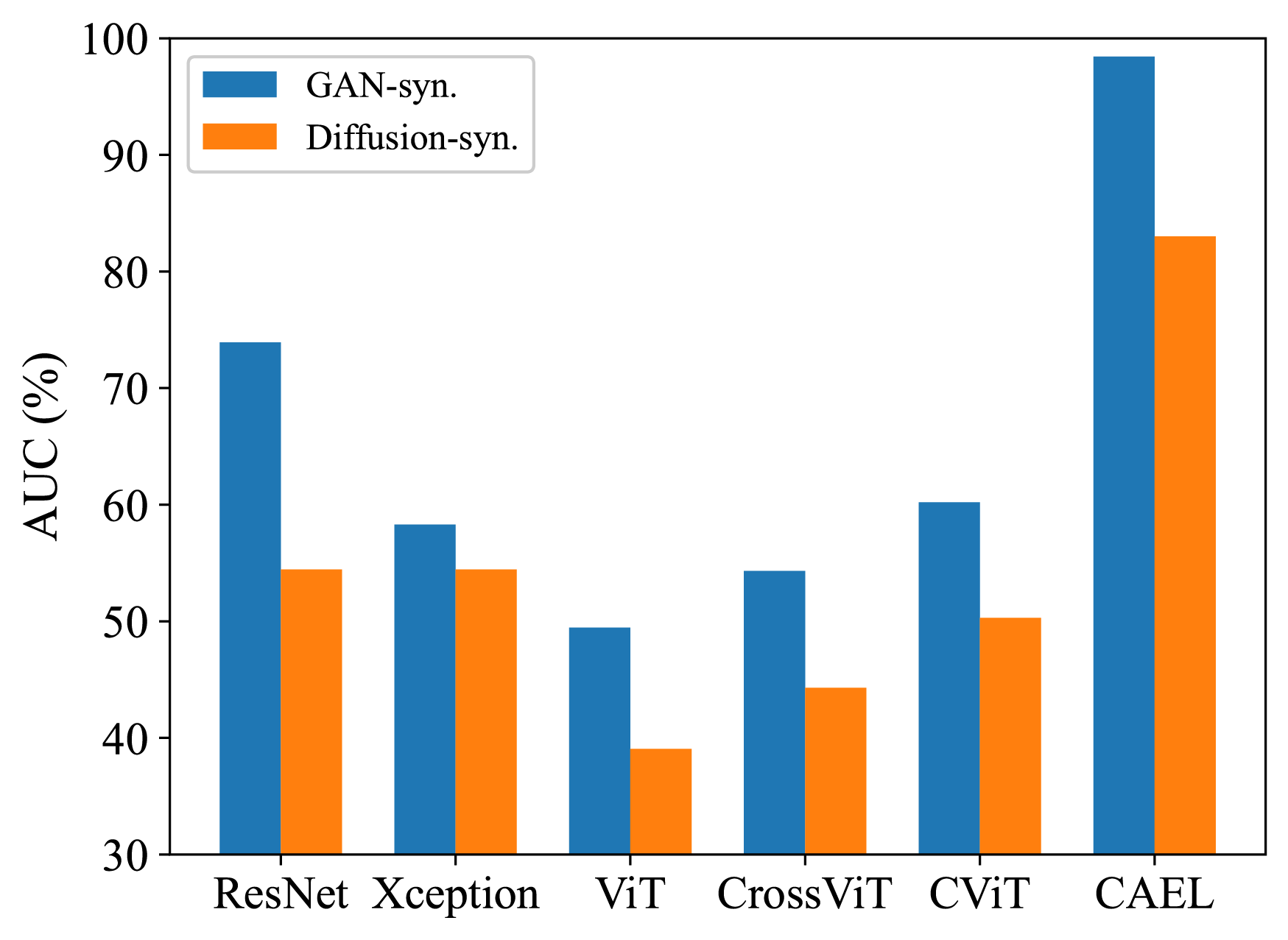}
	\centering		
	\caption{Comparison of the performance of the model on diffusion-based and GAN-based generators. Models are trained on diffusion-based generators and tested on GAN-based ones, and vice versa.}\label{fig17}
	
\end{figure}

\section{Conclusion}\label{sec6}

In this paper, we establish a large-scale, diverse, and fine-grained high-fidelity deepfake dataset, namely GenFace, along with a robust deepfake detector. Our GenFace dataset contains novel forgery methods such as the diffusion-based generator, achieving diversity, high image quality, and hierarchical fine-grained classification, which benefits developing more robust deepfake detection techniques. Moreover, we propose the cross appearance-edge learning model to capture rich multi-grained appearance and edge global fusion forgery traces. 

{\bfseries\setlength\parindent{0em} Limitations.} Although we have conducted the heatmap visualization using Grad-CAM, to help CAEL achieve manipulation localization, we do not provide a mechanism for localizing manipulated regions. In the future, we propose to devote ourselves to face forgery localization and detection tasks. Firstly, we would provide forgery masks as ground truth, and design a lightweight face forgery localization module as a side branch, to augment manipulation localization capabilities of detectors. Secondly, face forgery detection and localization can be regarded as a multi-task learning scenario, where the model simultaneously learns both detection and localization by shared representations. To furgher improve the interpretability and explainability of the detector, a detailed workflow can be provided using the formal formula, to highlight the key components and how they contribute to the overall performance.

Despite of the better diversity than other face forgery datasets, GenFace still needs to incorporate more abundant facial images, to further enhance the robustness and comprehensiveness of our dataset. In the future, we plan to introduce facial images generated by additional advanced generative models such as variational autoencoders (VAEs) and normalizing flow models, to facilitate detectors to mine a wider range of forgery traces, ultimately contributing to the advancement of deepfake detection.
\bibliographystyle{IEEEtran}
\bibliography{cmsevit_bib}

\end{document}